# The use of GPT-4o and Other Large Language Models for the Improvement and Design of Self-Assessment Scales for Measurement of Interpersonal Communication Skills

**Goran Bubaš**

University of Zagreb

Faculty of Organization and Informatics

Pavlinska 2, 42000 Varaždin, Croatia

`gbubas@foi.unizg.hr`

**Abstract.** *OpenAI's ChatGPT (GPT-4 and GPT-4o) and other Large Language Models (LLMs) like Microsoft's Copilot, Google's Gemini 1.5 Pro, and Antrophic's Claude 3.5 Sonnet can be effectively used in various phases of scientific research. Their performance in diverse verbal tasks and reasoning is close to or above the average human level and rapidly increasing, providing those models with a capacity that resembles a relatively high level of theory of mind. The current ability of LLMs to process information about human psychology and communication creates an opportunity for their scientific use in the fields of personality psychology and interpersonal communication skills. This article illustrates the possible uses of GPT-4o and other advanced LLMs for typical tasks in designing self-assessment scales for interpersonal communication skills measurement like the selection and improvement of scale items and evaluation of content validity of scales. The potential for automated item generation and application is illustrated as well. The case study examples are accompanied by prompts for LLMs that can be useful for these purposes. Finally, a summary is provided of the potential benefits of using LLMs in the process of evaluation, design, and improvement of interpersonal communication skills self-assessment scales.*

## 1 Introduction

After it was introduced in November 2022 by the OpenAI research organization (OpenAI, 2022), *ChatGPT* (GPT-3.5) and its successors like *GPT-4, GPT-4 Turbo*, and *GPT-4o*, as well as other *Large Language Models* (LLMs), including Microsoft's *Copilot* (formerly *Bing Chat*) and Google's *Gemini* (formerly *Bard*), rapidly attracted much interest by the research and academic community considering how they *compare to human subjects* in university exams, aptitude, and intelligence tests, as well as complex cognitive problems. For instance, Kocoń et al. (2023), among other authors, thoroughly investigated the capabilities of GPT-4 in several well-known *natural language processing* (NLP) tasks that demonstrate analytical skills in language analysis and understanding, for instance regarding simple text classification (spam, humor, sarcasm, etc.), sentiment analysis and emotion recognition, reasoning within a personal context, semantic annotation like word sense disambiguation and other. They concluded that GPT-4 solved most of such problems considerably well and also exhibited applicability to generally different tasks. Furthermore, GPT-4 manifested the capacity to explain its output but did not perform, as well as the best available specialized solutions for NLP. Such sophisticated capacities of the ChatGPT family of LLMs rapidly attracted interest for their scientific research application in computer science, healthcare, language studies, education, psychology, and other fields.

Among various artificial intelligence (AI) tools that could assist academic researchers and educators, ChatGPT offered particular benefits owing to its diverse capacities (for instance, see: Borger et al., 2023). In a review paper, Chukwuere (2024) listed several broad areas in which ChatGPT was able to assist in research, including (a) academic research writing, (b) data analysis, (c) literature review, and (d) scientific collaboration. As an example, in the specific area of *STEM motivation* research, Donmez et al. (2023) provided a case study on how ChatGPT could assist in the following steps of designing and conducting research: defining research questions, refining the title and abstract, research design (e.g. method, sample, instruments, data analysis) and interpretation of the results of empirical research. Similarly, in their case study on another specific research topic of *the adoption of AI in higher education* Rahman et al. (2023) have illustrated that ChatGPT (i.e. GPT-3.5) can be utilized for writing the sections of a demo research manuscript that include: (a) research idea generation, (b) background/introduction, (c) problem statement, (d) literature review references, (e) methodology, (f) results and discussion. Furthermore, Khlaif et al. (2023) demonstrated that ChatGPT can generate 'original' manuscripts in the specific field of *teacher technostress* that mimic scientific research with the potential for positive evaluation by reviewers. Finally, in their case study of actual *peer reviews for papers that were submitted to four scientific conferences*, Liang et al. (2023) revealed that "between 6.5% and 16.9% of text submitted as peer reviews to the four observed conferences could have been substantially modified by LLMs", indicating the potential use of LLMs for peer-reviewing.

According to Rice et al. (2024), ChatGPT can be viewed as a "linguistic calculator" that can assist researchers in more efficient retrieval of information and design of experiments, even though its use also implicates the need for adherence to rigorous ethical standards. This was especially elaborated by Resnik and Hosseini (2024), who provided specific recommendations for researchers regarding, among other issues, their responsibility for AI-related biases and errors, disclosure of how they used AI in their research, and specifying the parts of the research output (including text and data) that were generated by AI. It must be highlighted that years before the introduction of ChatGPT various other AI tools have been used for *literature review* (Bolanos et al., 2024) and that those based on LLMs like ChatGPT are only a recent novelty for researchers who wish to facilitate this part of their research process. Still, it must be emphasized that LLMs may not always perform well in academic research-related tasks, as was revealed in a recent case study for the specific field of *psychology* (Lehr et al., 2024). This study indicated that GPT-3.5 and GPT-4 may perform poorly in the roles of a *research librarian*, *data generator*, and *novel data predictor*, but, ironically, can operate moderately well in the role of a *research ethicist*. Nevertheless, the presence of LLMs in psychology applications has witnessed such rapid progress since the introduction of ChatGPT in November 2022 (see: Ke et al., 2024) that recommendations are being made for reviewers of psychological research papers on how to attend to important evaluation issues when examining related studies (Abdurahman et al., 2024b).

After the previous outline of possible uses of LLMs in academic research, in the following section of this paper the potential of GPT-4 and other LLMs for *processing textual content* (i.e. transforming, analyzing, and creating) will be briefly outlined. Then, a concise review of selected research on the applications of LLMs in *personality studies* will be presented. Next, in the central part of this paper, specific utilizations of LLM tools will be demonstrated and analyzed in the form of *brief case studies* aimed at assisting researchers and educators in *evaluating,*

*improving, and creating interpersonal communication skills self-assessment scales and questionnaires*. It must be noted that (a) self-assessment scales are commonly based on verbal content in textual form and that (b) processing diverse information in that form is one of the great capacities of advanced LLMs like GPT-4o, Gemini 1.5 Pro, and Claude 3.5 Sonnet, and similar.

# 2 Performance of LLMs in verbal and reasoning tasks

In March 2023 OpenAI provided a detailed overview of the advancement concerning the performance of ChatGPT in various university-level exams and aptitude tests, with a noticeable improvement that had been achieved from GPT-3.5 to GPT-4 (OpenAI, 2024). Most of the offered data indicated that, when academic and professional exams are concerned, GPT-4 predominantly scores in the range from the 80th to 100th percentile when compared to human test-takers. However, despite the rapid and continuous evolution of LLMs, it must also be observed that such favorable results were not reported for ChatGPT/GPT-4 in one review study published in 2024 (Newton and Xiromeriti, 2024).

The technical evolution of ChatGPT from GPT-1 to GPT-4 in the 2018 to 2023 period (a comprehensive overview is obtainable in Wu et al., 2023) has been paralleled by a remarkable increase in its emergent abilities in *human-like reasoning* (see: Hagendorff et al., 2023). In that respect, evidence from peer-reviewed and published research includes one study that compared ChatGPT (version: January 9, 2023) to human subjects in *diverse problem-solving tasks* to find no substantial advantage for either (Orrù et al., 2023), and another study that found that GPT-3 can be similar or even above the level of human participants in *analogical reasoning tasks* (Webb et al., 2023). However, it is noticeable that, during the evolution of LLMs, in the tasks associated with intelligence testing with the use of verbal/textual items most new generations of LLMs have progressed in quantum leaps (for example, regarding the advancement from GPT-3.5 to GPT-4; see: McIntosh et al., 2024). Several articles *without scholarly peer review* have indicated that LLMs have the potential to surpass most human subjects in specific intelligence tests: King (2023), Klein & Kovacs (2023), Roivainen (2023), de Souza et al. (2023), Wasilewski & Jablonski (2024). Furthermore, according to *published scholarly papers with peer review*, recent research has demonstrated the potential of LLMs like GPT-4 to exceed human subjects in specific *divergent thinking tasks* that are commonly associated with creativity, for instance, in the studies by Guzik et al. (2023) and Hubert et al. (2024). Despite potential *controversies*, outside the scholarly community there has been a considerable focus on the *issue of humans versus LLMs regarding* *cognitive abilities* (for instance, see: Thompson, 2024), which has attracted much interest from the general public. On the other hand, it could be theorized that verbal information (test items) for cognitive-like tasks given to LLMs may not measure the ability that is biologically and psychologically comparable to the cognition of human subjects (for this perspective refer to: Heaven, 2023; Chapman, 2023).

It is the *manifested verbal and reasoning abilities* of LLMs, especially GPT-4 and GPT-4o, that motivated their *testing and evaluation in tasks related to the design and improvement of self-assessment scales and questionnaires in personality psychology*. For instance, Bubeck et al. (2023) illustrated that GPT-4 can solve novel and difficult tasks close to human-level performance in disciplines like mathematics, medicine, law, and psychology. Their demonstrations supported the assumption that verbal and reasoning abilities of GPT-4 could be placed into service by the authors of self-assessment scales. Specifically, regarding the area of psychology, Bubeck et al. concluded that GPT-4 can "infer what each character's mental state is, and also discern where miscommunication and misunderstanding lie" in a verbal description of their situation, as well as to "reason about the emotional state of others in complex situations" and "reason about the intentions of people in complex social situations", suggesting that "GPT-4 has a very advanced level of theory of mind".

# 3 LLMs and personality research

Even though the use of LLMs like ChatGPT (and other more advanced models) in various areas of personality research is only in the initial phases, besides the efforts to decipher the *personality of popular LLMs* (for a brief review of this topic see: Wen et al., 2024), several important areas of interest have also emerged that will be elaborated in the continuation of this section.

## 3.1 Assessment of personality traits

A considerable number of studies have addressed the ability of LLMs to *assess human personality*. For instance, Rhao et al. (2023) explored this issue with imaginary personalities and tried to propose a *general framework for personality assessment* by LLMs. A study performed by Cao and Kosinski (2024) with GPT-3, an earlier version of ChatGPT, revealed that the way human raters perceive public figures regarding *likability* and *Big Five personality traits* can be predicted by "their names' location in GPT-3's semantic space". Guinn (2023) discovered that in classifying human author's writings using the *Myers-Briggs Personality Type (MBPT) dimensions* (i.e. Introversion-Extroversion, Intuition–Sensing, Thinking–Feeling, Judging–Perceiving) the classifications of ChatGPT (GPT-3.5) were comparable to other state-of-the-art models that used *machine learning* for personality type classification.

The same ability of GPT-3.5 and GPT-4 to predict *MBPT personality dimensions* was confirmed by Murphy (2024) concerning the classification of personality type that was based on the 50 most recent tweets by an individual. Also, in the latter study, GPT-3.5 and GPT-4 substantially outperformed three machine learning models at the same task. Similarly, the research conducted by Peters and Matz (2024) also suggests that GPT-3.5 Turbo and GPT-4 display a modest ability to infer *Big Five personality traits* from social media posts of Facebook users without being previously trained for that purpose (in contrast to machine learning models), even though the accuracy of the LLMs in that study was slightly lower than that of models that were specifically trained for that purpose. Ji et al. (2024) revealed a rather high ability of GPT-3.5 Turbo to predict *HEXACO personality dimensions* (i.e. Honesty-Humility, Emotionality, Extraversion, Agreeableness, Conscientiousness, and Openness to Experience) from basic persona descriptions. In addition, Ji et al. (2023) explored the ability of ChatGPT for personality recognition from human text (essays and tweets) and found that it was capable of providing predictions of *Big Five personality traits* (i.e. Openness, Conscientiousness, Extraversion, Agreeableness, and Neuroticism) that are comparable to other automated models, while also supplementing these predictions with natural language explanations and logical reasoning. However, the study performed by Derner et al. (2023) uncovered only rather low correlations between the self-assessment of subjects on the one side, and *Big Five personality* predictions made by GPT 3.5 from their short texts, on the other.

## 3.2 Desirability rating of personality items

Phillips and Robie (2024) investigated the efficacy of utilizing LLMs to *identify ideal responses* in personality assessments with the use of *integrity* and *conscientiousness* measures and revealed that GPT-3.5 and GPT-4 performed better than other LLMs in that respect, as well as that they outperformed human participants (university business students) in the ability to fake responses to score highly on favorable personality traits in a job application and selection situation. It must be noted that the LLMs specially trained for *sentiment analysis* (other than those of the ChatGPT type) have proven comparable to human judges in *desirability rating of items* of self-report measures (Hommel, 2023), but are far more complicated for such use by researchers than the currently available advanced models like GPT-4o, Gemini 1.5 Pro, and Claude 3.5 Sonnet.

## 3.3 Emotion classification and understanding

Belkhir and Sadat (2023) revealed a rather high accuracy of ChatGPT in *emotion classification*, as well as the possibility to pre-train ChatGPT for the production of more human-like responses in empathetic conversations. The research performed by Welivita and Pu (2024) suggests that the *verbal responses to emotional situations* created by GPT-4 after specific prompting can exceed those created by humans. There is an indication that ChatGPT (GPT-4) could even surpass physicians in *delivering empathetic communication* in interaction with real-world patients (Luo et al., 2024). In a study conducted by Elyoseph et al. (2023), ChatGPT demonstrated a high level of *emotional awareness* (i.e. cognitive ability that enables an individual to conceptualize their own and others' emotions) that was considerably above the average level of a large sample of male and female human subjects. As many as 18 LLMs (including 10 OpenAI versions of ChatGPT) were tested by Wang et al. (2023) regarding *emotion understanding* (as a component of the "emotion quotient"). In this study, GPT-4 surpassed all other LLMs with a score that was also superior to 89% of the human subjects who had taken the specially constructed *Emotion Understanding* test. Similar results were obtained in research conducted by Brin et al. (2023) regarding GPT-4 using the selected items (related to *communication skills, ethics, empathy, and professionalism*) of the *United States Medical Licensing Examination*. Again, in the latter study, GPT-4 surpassed human subjects and demonstrated potential in "handling complex ethical dilemmas that involve empathy which is critical for patient management by physicians". Xu et al. (2023) revealed that LLMs perform better in *academic intelligence* measures than in situational *social intelligence* and that, among several LLMs, GPT-4 performed best in most *academic intelligence* measures, as well as second best (after GPT-3.5 Turbo) on a measure of *social intelligence*.

### 3.4 Psychiatric disorders

Finally, in the field of *psychiatry* (Abbas et al., 2023) there are indications that ChatGPT and Google Bard can identify various disorders (e.g. *major depressive disorder, schizophrenia, obsessive-compulsive disorder*, etc.) from case descriptions made by a physician based on the Diagnostic and Statistical Manual of Mental Disorders (DSM). Also, in a study by Hadar-Shoval et al. (2024), ChatGPT demonstrated mentalizing-like abilities regarding the feelings of imaginary individuals who had been attributed to possessing *borderline personality disorder* or/and *schizoid personality disorder* in various simulated situations that were designed according to the *Levels of Emotional Awareness Scale*.

### 3.5 Item generation for measurement of personality traits

The previous outline of selected topics related to the use of ChatGPT and other LLMs in *personality research* provides evidence for their capacity to (a) 'reason' about human personality traits with a considerable level of 'comprehension'; (b) based on textual input, 'detect' or 'identify' specific personality traits, like those measured with Big Five and MBPT questionnaires; (c) depending on the tasks at hand, surpass humans and/or machine learning models in personality classification tasks; (d) classify and 'manifest' selected emotions, (e) display 'empathy', 'emotional awareness' and 'emotion understanding', as well as (f) show the 'ability' to identify specific mental disorders and 'mentalize' how some disorders would impact individuals in simulated situations. These findings provide justification for the *main goal* of the research presented in this paper, which is to investigate and demonstrate the use of LLMs like GPT-4o for the evaluation, improvement, and design of self-assessment scales for the measurement of interpersonal communication skills. Having that context in mind, an overview of the use of LLMs for generating items for measures of personality traits is provided.

Most of the research on *automated item generation* (AIG) had utilized machine learning algorithms and neural networks, thus being more experimentally than practically oriented (for a brief review see: Circi et al., 2023), with numerous studies published before the release of LLMs like OpenAI's *ChatGPT* in 2022 and Google's *Bard* (later named *Gemini*) in 2023. However, only several early AIG studies were related to personality assessment. For instance, in one of the pre-ChatGPT studies, von Davier (2018) used a deep learning neural network approach and probabilistic language models to generate Big Five personality items, but with mixed results. The potential methods and challenges for pre-ChatGPT AIG of personality items were described by Hommel et al. (2022), who fine-tuned a pre-trained GPT-2 based AIG model using selected 1715 unique item stems from the *International Personality Item Pool* that were grouped by associated construct labels. Their *LLM-based method* generated a large pool of items from which personality assessment scales were formed and administered online to human subjects (N=220) to evaluate them and compare them with *human-authored* items and scales. The results of their empirical evaluation indicated that machine-authored items for personality assessment were not "comparable in structural validity to what should be expected from a psychometrically sound personality questionnaire". Using the same type of pre-trained GPT-2 model, Götz et al. (2023) developed their *Psychometric Item Generator*, i.e. a natural language processing algorithm that could produce pools of items for novel constructs (e.g. a newly imagined "wonderlust" scale), as well as scales that measure traditional constructs like those of the Big Five model of personality. These authors have also performed rigorous testing of their *machine-generated* measure of Big Five traits using two large samples of human subjects ($N_1=501$, $N_2=773$) and revealed that it is psychometrically correspondent (regarding reliability, content, structural and criterion validity) to a popular *human-authored* short form of the Big Five test (i.e. BFI-2, see: Gosling et al., 2007).

To generate Big Five personality items for self-assessment scales, Lee et al. (2023) used a GPT-3 pre-trained model, which was approximately 100 times larger than GPT-2, enabling them to use only a "prompt-based few-shot learning approach" with labeled examples. After a thorough selection by human evaluators of items, the final machine-generated Big Five personality test consisted of 25 items that they administered online (together with a human-generated Big Five test) to 420 subjects. The results of the confirmatory factor analysis suggested that the *machine-authored* test adequately represented the Big Five personality structure and that it was comparable to the *human-authored* Big Five test not only in that personality domain but also regarding the McDonald's omega (ω) measure of reliability and several criterion-related validities. This study by Lee et al. (2023) demonstrated that in parallel to the advancement of LLMs, the sophistication and diversity in the approaches to developing machine-authored items for personality tests have also evolved.

The potential of ChatGPT to generate, *with a simple single prompt*, 'ready to use' short scales (in Likert-type forced response choice form) to measure Big Five traits like conscientiousness, agreeableness, and emotional stability, comparable to those created by human experts, was demonstrated in a small-scale study by Walton and Anguiano-Carrasco (2024). Finally, using GPT-4 for generating and testing examples of prompts, Franco-Martínez et al. (2023) illustrated the possibilities to, among other tasks, use an LLM for the following: (a) generating items for measurement of non-cognitive (e.g. personality) constructs, including parallel items and scales; (b) designing non-cognitive items using an LLM in different item forms (i.e. Likert-type, forced choice, situational judgment, semantic differential), for different populations and languages; (c) analysis and improvement of item wording of non-cognitive items and tests, including content evaluation, construct representation and internal structure; (d) acting as a virtual respondent to non-cognitive items; (e) assisting in data analysis in form of advice; (f) performing data analysis and interpretation of results; (g) creating programming code, for instance using R script, for specific types of data analyses. Having in mind the previously outlined developments in using LLMs like GPT-3 and GPT-4 for research in personality assessment, including item generation for scales and questionnaires that measure personality traits, the possibility to use such tools for generating, evaluating and improving *interpersonal communication skills questionnaires and scales* with advanced LLMs (i.e. GPT-4o, GPT-4, Copilot, Gemini 1.5 Pro, Claude 3.5 Sonnet) will be analyzed in the following sections of this paper.

# 4 Goals and research questions

## 4.1 Research goal

The main goal of the study that is presented in this paper was to *explore and provide examples of several means by which LLMs like GPT-4o, GPT-4, Copilot, Gemini 1.5 Pro, and Claude 3.5 Sonnet can be used for the improvement, evaluation, and design/construction of self-assessment scales in the scientific field of interpersonal communication skills*. Because of the limited length of this article, only brief demonstrations and examples of potential uses of such tools will be provided. At the time this article was being completed *no scholarly articles were found that extensively addressed the issue of the improvement and design of self-assessment scales for the measurement of interpersonal communication skills*.

## 4.2 Research questions

The following research questions were formulated based on the previous *research goal* and potential benefits for scholars and educators who are involved in the process of design and evaluation of self-assessment scales in the specific field of *interpersonal communication skills*:

*RQ1:* Can Microsoft Copilot and other LLMs be effectively used for the translation and simplification of the items of self-assessment scales?

*RQ2:* Can prompts be used with GPT-4o and other advanced LLMs to categorize the items of questionnaires with multiple self-assessment scales in a form that complements their empirical evaluation using confirmatory factor analysis?

*RQ3:* Can GPT-4o and other advanced LLMs be effectively utilized in activities like evaluation, improvement, and redesign of interpersonal communication skills self-assessment scales and questionnaires?

*RQ4:* Can GPT-4o and other advanced LLMs generate/design new interpersonal communication skills self-assessment scales and questionnaires?

*RQ5:* "Can GPT-4o, Gemini 1.5 Pro, and Claude 3.5 Sonnet generate new scales and respective items for the measurement of interpersonal communication skills, and subsequently apply them to users in an interactive multiturn form?"

# 5 Method

## 5.1 Instruments

As was previously outlined, LLMs of the ChatGPT family have considerably improved in their capacity for semantic processing of textual/verbal information and also demonstrate considerable *explicit* and *implicit* knowledge in the field of human personality. Various providers of AI-based tools and services, like OpenAI, Microsoft, Google, Antrophic, and others, are competing in creating newer, increasingly more sophisticated, technically proficient, and advanced releases of their products, currently at a rate of at least once or twice a year. In conducting a research study involving LLMs, it is opportune to consider (a) currently up-to-date LLMs, as well as (b) those that are available to other researchers. In addition, the training data (i.e. digitalized human knowledge in textual form) for LLMs like GPT-4/GPT-4o, Copilot, Gemini 1.5 Pro, and Claude 3.5 Sonnet is constantly expanding and includes a vast amount of content that is available on the internet. That means that at least some responses to the prompts delivered to LLMs trained on internet data and with real-time access to internet content may not only be the result of their capacity for semantic analysis and 'transformation' of text, but also of some form of ability to retrieve content from their 'memory', or as a result of concurrent internet search (as in the case of Microsoft's Copilot, for example). Furthermore, to select the currently best-performing LLMs it is recommended to consult LLM benchmarks and leaderboards (Caballar and Stryker, 2024). However, the problem remains regarding the potential LLMs' responses that are generated from 'memory' or internet search which may represent a 'copy' or highly resemble a solution that can be found elsewhere in an (almost) identical form, as illustrated by the following example.

To test the ability of Microsoft's Copilot to sort usability and user experience-related items from an assessment instrument published in an open-access online scholarly journal research paper (see: Bubaš et al., 2024) in December 2023, the following prompt was given to Copilot in June 2024: "Categorize the following items/sentences to the usability categories/variables 'Perceived Usefulness', 'General Usability', 'Learnability' and 'System Reliability' <followed by a list of 19 items:>". Surprisingly or not, Copilot performed this categorization perfectly, with an identical order of items as in the original instrument that had been published online, also returning a link to this research paper. This indicates that in testing various 'abilities' of LLMs regarding the design and evaluation of items and scales for self-assessment, it may sometimes be opportune to use unpublished assessment instruments that could not have been a part of an LLMs 'training' or easily available to them by internet search.

### 5.1.1 Large Language Models (LLMs)

Having in mind the results of the LLM benchmarking in June 2024, as well as their other characteristics like availability, popularity, and potential for parallel use on equal tasks, the following LLMs were selected for use in case studies that are presented in this paper: GPT-4o and GPT-4 (OpenAI), Copilot (GPT-4 Turbo, Microsoft), Gemini 1.5 Pro (Google) and Claude 3.5 Sonnet (Antrophic).

The preferred LLM for the translation of questionnaire items from Croatian to English and their simplification (in *section 6.1*) was *Copilot* (powered by GPT-4 Turbo).

Initial categorizations of scale items were performed with *GPT-4o* and sometimes replicated with the use of other LLMs.

Prompts about different interpretations, aspects of use, and analyses of assessment scales and items were delivered primarily via *GPT-4o* and occasionally also tested with other LLMs (*GPT-4, Copilot, Gemini 1.5 Pro*, and *Claude 3.5 Sonnet*). In other words, OpenAI's GPT-4, Microsoft's Copilot, Google's Gemini 1.5 Pro, and Antrophic's Claude 3.5 Sonnet were used in parallel or for comparison with GPT-4o.

The author would like to emphasize that the text of this article was fully written by the author (without the use of AI). The final text of the article was proofread by a human expert (with a Ph.D. in *English studies*) for translating and editing IT-related texts.

### 5.1.2 Self-assessment scales and questionnaires

For case studies presented in this paper, several unpublished questionnaires and self-assessment scales for the measurement of interpersonal communication skills were used which are described below.

The *Interpersonal Communication Competence Inventory (ICCI)* (Bubaš, 2003) is a multifaceted questionnaire designed only for research purposes with acceptable internal consistency of all self-assessment scales and concurrent validity regarding other interpersonal skills measures and the Big Five (NEO-PI) personality traits. The ICCI consists of 374 items distributed in 23 scales that are designed to measure the following constructs: *Knowledge of the Communication Process, Motivation for Interpersonal Communication, Social Relaxation, Composure, Initiation of Interaction, Communication Effectiveness, Assertiveness, Interpersonal Control, Interaction Management, Adaptability, Interaction Involvement, Self-Monitoring, Decoding Skills, Nonverbal Sensitivity, Encoding Skills, Nonverbal Expressivity, Verbal Expressivity, Altercentrism / Other-orientedness, Self-Disclosure, Empathy, Support, Collaboration*, and *Conversational Skill*. The English versions of the *selected sample items* for all the scales mentioned above are provided in *Appendix I*. The ICCI self-assessment scales were designed in 1999 (in Croatian language) based on the literature on related constructs. Since the items of the ICCI scales were never published in their Croatian or English version they can be used for *zero-shot* and *one-shot* prompting of LLMs that were not previously trained with the specific input regarding the content of the ICCI. It must be noted that almost all of the ICCI scales comprise exactly 8 positively worded and 8 negatively worded (reverse-scored) items each (except for the *Motivation for Interpersonal Communication* scale, which has 15 positively and 15 negatively worded items, and the *Verbal Expressivity* scale, which has only 8 positively worded items).

The *four shortened versions of ICCI scales* (with 4 positively worded items each) that measure the following constructs (abbreviation and respective Cronbach alpha coefficients are in brackets) comprise *Verbal Expressivity* (VE; α=.77), *Self-Disclosure* (SD; α=.70), *Composure* (CO; α=.75), and *Conversational Skill* (CS; α=.81). These short forms of selected interpersonal communication skills scales were created in year 2022 by adapting the selected items of the corresponding ICCI scales in the Croatian language for a pilot study on communication skills in using English as a foreign language (Bubaš and Kovačić, 2023). For demonstration purposes (see *Table 1*), the items of these four scales were translated into English using Microsoft's Copilot and, without alteration, further used in this English form with other LLMs in the case studies presented in this paper. The English versions of all 16 items are displayed in *Table 2*. For their response to the items of the aforementioned scales, the subjects were asked to rate how much the content of each item is characteristic of themselves (the Likert-type response scale was "1 = Very little", "2 = Little", "3 – Average", "4 = Much", "5 – Very much"). Because of the brevity of these four scales and the relatively high diversity (semantic 'distance') of constructs that they measure, they can be used for research and demonstration purposes with different LLMs. As in the case of ICCI, the items of the four brief scales were not earlier published, so they can be used for *zero-shot* and *one-shot* prompting of LLMs that were not previously trained with their specific content.

The **Active Listening** self-assessment scale was developed by the author of this study in year 1995 exclusively based on related literature. It comprises 25 heterogeneous items reflecting *various aspects of active listening* as an interpersonal communication skill. It has been used in Croatian and English forms for education (in academic teaching and training seminars). The Croatian and English versions of this scale were not previously published. The items of the *Active Listening* scale that were translated to English by a non-expert are listed in *Appendix III*.

## 5.2 Subjects

For the empirical evaluation in form of confirmatory factor analysis (see *Table 2*) of four shortened versions of interpersonal communication skills scales (labeled *Verbal Expressivity*, *Self-Disclosure*, *Composure*, and *Conversational Skill*) the data was collected in the year 2022 and the subjects were first-year undergraduate students of information systems at a Croatian university (N=170; 17 to 23 years of age; 77.7% male, 19.4% female and 2.9% that did not disclose gender).

The data presented in *Appendix II* was collected using the ICCI questionnaire in 1999 (N=403). The subjects were first-year Information Systems students from the University of Zagreb in Croatia, of whom 72% were male and 28% were female.

## 5.3 Procedure

For each case study a different procedure was used to respond to the main goal of this study and research questions.

# 6 Case studies

Each of the following case studies was designed to answer one of the research questions. Using the same or similar prompts for GPT-4o and other LLMs as in the case studies researchers can replicate most of the processes in those case studies as long as the examined LLMs are available. With the ongoing advancement of LLMs novel AI products and services may considerably outperform the ones used in this research. However, most of the principles and techniques for the evaluation, improvement, and design of self-assessment scales and their items with the use of examined LLMs GPT-4/GPT-4o, Copilot, Gemini 1.5 Pro, and Claude 3.5 Sonnet, will probably also continue to be applicable with the appearance of newer and better versions of such tools/services or new LLMs like *Llama* (Meta AI) or *Grok* (xAI). Part of the focus in the presentation of the following case studies will be on the *formulation and testing of useful prompts* that can be utilized in the creation and evaluation of items of self-assessment scales and questionnaires in the field of interpersonal communication skills.

## 6.1 Use of LLMs for the translation and simplification of self-assessment scale items

One of the important aspects of the design and use of novel self-assessment scales is the possibility of translating them to another language to present them to an international scholarly audience and/or apply them (online) to subjects of different native language(s) and geographical location(s). Also, the authors of self-assessment scales may wish to phrase the wording of items in initial versions of their scales in a way that is simpler and easier to understand by the subjects in their research. *Table 1* shows the translation by Microsoft's Copilot (in the middle column) into English of items from two previously unpublished interpersonal communication competence scales in the Croatian language (in the left-hand column) entitled (a) *Verbal Expressivity* and (b) *Self-Disclosure*. The translated English versions of the items were also used to demonstrate the simplification of those items (in the right-hand column). It has to be noted that the two scales represent a shortened (i.e. four-item) version of the original scales developed by the author of this paper, as explained earlier in *subsection 5.1.2*. The initial prompt for the translation task was: "**Translate the following sentences from Croatian to English. These sentences are from a self-assessment scale related to self-disclosure [verbal expressivity] in interpersonal communication: <a list of items>**". After the translation is completed, the prompt for the simplification task can, for instance, be worded in the following form: "**Now try to rephrase the following sentences to make them easy to read and understand** as possible by common folks, but still retain their meaning and that they represent the interpersonal skill of self-disclosure [verbal expressivity]: <a list of items>".

It must be noted that the attempted replications of this simplification task are likely to result in somewhat different wording of the sample items in *Table 1* (but with basically equivalent meaning) depending on the previous context of a conversation with Copilot and potential slight changes in the wording of the prompt (for instance, this happens if the wording of the initial prompt above is changed to "Simplify the sentences below so that they are easy for everyday people to understand, while still conveying the idea that they demonstrate the interpersonal communication skill self-disclosure [verbal expressivity]: <a list of items>").

Researchers may choose to ask different LLMs to perform the same translations or simplifications and select the best result for each item. In fact, all of the examined LLMs (i.e. GPT-4, GPT-4o, Copilot/GPT-4 Turbo, Gemini 1.5 Pro, and Claude 3.5 Sonnet) performed comparably well on the translation task exhibited in *Table 1*, as well as in the *back-translation* of the items from English to Croatian.

**Table 1.** Example of the translation and simplification of the wording of the items of the shortened versions of the *Verbal Expressivity* and *Self-Disclosure* self-assessment scales (Microsoft's Copilot, June 2024; the number of characters with spaces for each item in the English language is given in brackets)

| ORIGINAL ITEMS (CROATIAN) | TRANSLATION BY COPILOT | SIMPLIFICATION BY COPILOT |
|---|---|---|
| **Verbal Expressivity scale** | | |
| Dobro mogu animirati druge prepričavanjem raznih zanimljivih događaja i anegdota. | I can easily engage others by recounting various interesting events and anecdotes. (82) | I can capture people's attention by sharing interesting stories and events. (75) |
| Moje je izražavanje bogato dojmljivim paralelama, metaforama, primjerima i slikama. | My expression is rich with impressive parallels, metaphors, examples, and images. (81) | When I talk, I use vivid comparisons, metaphors, and examples. (62) |
| Lako mi je riječima opisati nešto poput pejsaža u prirodi, slike ili glazbene kompozicije. | I can easily describe something in words, such as a natural landscape, a picture, or a musical composition. (108) | I'm good at describing things using words, like landscapes, pictures, or music. (79) |
| Čak i manje uzbudljive pojave ili događaje opisat ću drugima na zanimljiv i privlačan način. | Even less exciting occurrences or events, I will describe to others in an interesting and appealing manner. (107) | Even mundane stuff becomes interesting when I talk about it. (60) |
| **Self-Disclosure scale** | | |
| Mogu dobro procijeniti kojim osobama i u kojoj situaciji mogu izjaviti povjerljive činjenice o sebi. | I can accurately assess which individuals and in which situations I can disclose confidential facts about myself. (113) | I know when it's okay to share personal things about myself with different people and in various situations. (108) |
| Izborom osobnih misli i osjećaja o kojima informiram druge djelujem im privlačnije i stječem njihovu naklonost. | By selecting personal thoughts and feelings that I share with others, I appear more attractive to them and gain their favor. (124) | When I talk about my thoughts and feelings, it makes me more appealing to others and they like me more. (103) |
| Vodim računa o tome da drugima priopćavam one povjerljive i osobne informacije o meni koje su im prihvatljive. | I take care to communicate with others the confidential and personal information about myself that they find acceptable. (121) | I'm careful to share only the personal information that others are comfortable hearing. (87) |
| Kad izjavljujem o sebi nešto vrlo osobno i privatno, uspijevam postići veću bliskost s drugim osobama i ne udaljiti ih. | When I express something very personal and private about myself, I succeed in achieving greater closeness with other individuals and not distancing them. (153) | When I open up about something private, it helps me get closer to others instead of pushing them away. (102) |

Using one LLM for the translation of items from a specific language (in this study, from Croatian) into English and another LLM for the translation of the same items from English to the first/original language (in this study, into Croatian) can be regarded as *back-translation*, e.g. a quality check procedure in using scales for different first language groups of subjects, international research cooperation, and scholarly publishing. It has been claimed that, while back-translation should be a common practice in psychological research, additional techniques beyond traditional back-translation can also be used to examine equivalence in more detail (Klotz et al., 2023). Quality *forward* and *backward* translation ensures that self-assessment scales keep the psychometric qualities of the original scale in another language, i.e. that the object of measurement is not diluted or altered by translation (Fenn & George, 2020).

The example exhibited in *Table 1* illustrates that Microsoft's Copilot (as well as LLMs with comparable or better translation capability, see: Wiik, 2024) can be effectively used for:

• Initial translations of items in self-assessment scales from Croatian to English;

• Simplification of the wording of the items in the English language and shortening of sentences without substantial changes in their meaning (each of the *simplified* sentences of the 16 items from the 4 self-assessment scales was on average 27 characters *shorter* than their first translation into English performed by Copilot).

To further examine the potential of Copilot to simplify the items of self-assessment scales, the previously discussed prompt **"Try to rephrase the following sentences to make them as easy to read and understand as possible by common folks, but still retain their meaning and that they represent the interpersonal skill of <name of skill>"** was used with a much larger number of self-assessment scales and items of the English version of the *Interpersonal Communication Competence Inventory* that was designed by the author (Bubaš, 2003) for scientific measurement of the following 23 interpersonal skills and traits (the number of items in each scale is given in brackets): *Knowledge of the Communication Process* (16), *Motivation for Interpersonal Communication* (30), *Initiation of Interaction* (16), *Communication Effectiveness* (16), *Assertiveness* (16), *Interpersonal Control* (16), *Interaction Management* (16), *Decoding Skill* (16), *Nonverbal Sensitivity* (16), *Adaptability* (16), *Altercentrism / Other-orientedness* (16), *Composure* (16), *Self-Disclosure* (16), *Empathy* (16), *Comforting* (16), *Verbal Encoding Skills* (16), *Nonverbal Expressivity* (16), *Verbal Expressivity* (8), *Interaction Involvement* (16), *Cooperation* (16), *Social Relaxation* (16), *Self-Monitoring* (16), and *Conversational Skill* (16). The simplification using the previous prompt resulted in the shortening of all the 374 items of the ICCI. In fact, the total number of words and characters (with spaces) for the 374 items, depending on the wording of different prompts, could be in most cases reduced by more than 30% without substantial loss regarding the meaning of the original ICCI item (i.e. sentences) in English.

It must be noted that, by repeating the prompt regarding the simplification of the items (i.e. their translations by Copilot in the middle column of *Table 1*) of the **short versions** of the *Verbal Expressivity* scale and the *Self-Disclosure* scale, similar results (although with slightly different wording of simplified items) were obtained from GPT-4o, GPT-4, Gemini 1.5 Pro, and Claude 3.5 Sonnet. To conclude, the previously presented examples of the use of Copilot (for other LLMs the results are not shown here for brevity reasons) provide a *positive response to the first research question*: "RQ1: Can Microsoft Copilot and other LLMs be effectively used for the translation and simplification of the items of self-assessment scales?". However, it is the opinion of the author of this paper that the use of LLMs for the translation and simplification of scale items, although convenient for researchers, should by no means eliminate the need for proofreading of the final versions of the items and respective scales by a native language expert (in this case, a native speaker of English), hopefully with sufficient domain knowledge for which the items and scales are constructed.

## 6.2 Use of LLMs for the categorization of items of self-assessment scales

In scale development, the confirmatory factor analysis is often used in the *construct validation process* to investigate the latent structure of an assessment instrument (questionnaire), i.e. to verify that the number of its hypothetical underlying dimensions and loadings of items on specific factors are in accordance with the theoretical assumptions that guided its construction (see: Brown, 2015, pp. 1-2). The items from a newly constructed and *unpublished* instrument (Bubaš, 2023) for the self-assessment of four communication skills (*Verbal Expressivity, Self-Disclosure, Composure, and Conversational Skill*) will be used in this section to illustrate the use of LLMs for categorization of items in parallel to conducting *confirmatory factor analysis*. In *Table 2* the items and results of the confirmatory factor analysis are presented for the aforementioned four scales. Similar to the previous section of this paper, each of the scales represents a shortened (i.e. four-item) version of the original ICCI scales. The data were collected in the year 2022 with the Croatian versions of the scales from subjects (N=170) who were first-year university students of information systems aged 17 to 23 (average age was 19 years), 77.7% of whom were male and 19.4% of female gender (2.9% did not specify their gender).

It must be noted that even though this was a *forced factor analysis with Varimax rotation*, a four-factor latent structure was also indicated by the *Kaiser-Guttman rule* of preserving factors with an eigenvalue greater than one in the initial unrotated solution (having in mind the potential problems with such a criterion, as discussed in: Zwick & Velicer, 1986).

From the data exhibited in *Table 2* it can be concluded that for almost all the items of the four self-assessment scales (i.e. *Verbal Expressivity*, *Self-Disclosure*, *Composure*, and *Conversational Skill*), the confirmatory factor analysis indicates their predominant factor loading on the factor that is semantically aligned with the scale label. In that respect, the only exception is the item "SD2 – By selecting personal thoughts and feelings that I share with others, I appear more attractive to them and gain their favor", in which the predominant loading was not on the expected factor labeled *Self-Disclosure* (F2), but on the one labeled *Verbal Expressivity* (F1).

To investigate if GTP-4o will group the items of the four self-assessment scales into the same broad categories as was assumed during the questionnaire design (namely, into the categories that would correspond to the scale labels: Verbal Expressivity, Self-Disclosure, Composure, and Conversational Skill), the following prompt was used before the listing of 16 items that are shown in *Table 2*: "Categorize the 16 items related to interpersonal communication skills into four categories with four items in each category by placing the semantically most similar items in each category. With the categorization into four categories give each category a name according to the potential communication skill that the four items in a specific category could measure. Here are the 16 items: <VE1-VE4, SD1-SD4, CO1-CO4, CS1-CS4>". As can be observed from this prompt, GPT-4o was not given the names of categories but was told to make the groupings according to the semantic similarity of items and then name each category with a label that is a generalization in the form of the name of a communication skill that represents the semantic content of the four items in a specific category. Also, in the previous prompt, the items were grouped together as they appear in the scales (see *Table 2*), e.g. their sequence was not randomized.

**Table 2.** Results of confirmatory factor analysis for the questionnaire items of the self-assessment scales that measure *Verbal Expressivity* (VE), *Self-Disclosure* (SD), *Composure* (CO), and *Conversational Skill* (CS) (N=170; the items were administered in Croatian language; factor loadings below .40 are not displayed)

| ITEMS OF INTERPERSONAL COMMUNICATION SCALES<br>(*VE* – Verbal Expressivity, *SD* – Self-Disclosure, *CO* – Composure, *CS* – Conversational Skill) | F1<br>{Verbal Expressivity} | F2<br>{Self-Disclosure} | F3<br>{Composure} | F4<br>{Conversational Skill} |
|---|---|---|---|---|
| VE1 - I can easily engage others by recounting various interesting events and anecdotes. | .68 | | | |
| VE2 - My expression is rich with impressive parallels, metaphors, examples, and images. | .79 | | | |
| VE3 - I can easily describe something in words, such as a natural landscape, a picture, or a musical composition. | .63 | | | |
| VE4 - Even less exciting occurrences or events, I will describe to others in an interesting and appealing manner. | .74 | | | |
| SD1 - I can accurately assess which individuals and in which situations I can disclose confidential facts about myself. | | .71 | | |
| SD2 - By selecting personal thoughts and feelings that I share with others, I appear more attractive to them and gain their favor. | .60 | .43 | | |
| SD3 - I take care to communicate with others the confidential and personal information about myself that they find acceptable. | | .68 | | |
| SD4 - When I express something very personal and private about myself, I succeed in achieving greater closeness with other individuals and not distancing them. | | .76 | | |
| CO1 - I manage to make my communication never seem nervous or upset. | | | .75 | |
| CO2 - I can control my interaction with others so that it acts spontaneously and naturally. | | | .70 | |
| CO3 - I'm not constrained or inhibited when I need to have a dialogue with a stranger. | | | .74 | |
| CO4 - I can achieve that my emotions cannot be noticed in the messages I exchange with others. | | | .63 | |
| CS1 - I can change the roles of the speaker and the listener in the conversation unobtrusively and in harmony with the interlocutors. | | | | .76 |
| CS2 - I can redirect the topics of conversation with others so that everyone achieves their needs and goals. | | | | .76 |
| CS3 - In an informal conversation, I manage to stick to topics that are (almost) interesting to everyone. | | | | .72 |
| CS4 - I successfully harmonize the content and flow of the conversation with the opinions and feelings of the interlocutor. | | | | .63 |

In response to the previous prompt GPT-4o provided the same categorization of items as in the original scales, but with the following minor alterations in their labels:

- *Emotional Control in Communication* instead of *Composure;*
- *Conversational Flow Management* instead of *Conversation Skill;*
- *Expressive Storytelling* instead of *Verbal Expressivity;*
- *Self-Disclosure Management* instead of *Self-Disclosure.*

A close inspection of the names of categories provided by GPT-4o, as well as the semantic content of the items in each category indicates that the category labels generated by GPT-4o have at least some plausibility and represent an *alternative interpretation* for the constructs that are associated with the items within each scale. Similar output to the previous prompt, with slightly different category labels, was generated by GPT-4, Copilot, Gemini 1.5 Pro, and Claude 3.5 Sonnet. Obviously, the procedure shown in the previous example, where GPT-4o and other LLMs were asked to categorize given items, without receiving suggestions as to what those categories are, can support potential researchers in their critical analysis of not only the *items* that should measure and represent the theoretical constructs but also of the *constructs* themselves.

Since LLMs are more *probabilistic* than *deterministic* systems in their processing and output, they can return different results to the same or similar input or prompts. Therefore, the categorization of items from the four self-assessment scales was requested from GPT-4o again, with the same wording of the prompt, but with the items that were listed in a *randomized order* (i.e. CO1, CS3, CO4, VE3, VE1, SD2, CS4, SD4, SD3, CO2, SD1, VE2, CS2, CS1, CO3, VE4). Interestingly, as a consequence of the randomization of the list of items in the prompt, their categorization by GPT-4o and the labels given to the categories were somewhat different (a similar effect of equal randomization of items on derived category labels was subsequently also noticed with the use of LLMs like GPT-4 and Gemini 1.5 Pro). Here is the output from GPT-4o after the randomization of items (note: categorizations of items and names of the categories that differ from the previously presented output from GPT-4o are marked with an asterisk - "*"):

*Category 1: Emotional Control in Communication* (CO1, CO4, CO2, CO3).

*Category 2: *Engagement and Interest Management* (*CS3, *VE3, *VE4, *CS2).

*Category 3: *Expressiveness and Descriptive Skills* (VE3, VE2, *SD2, *SD4).

*Category 4: *Interpersonal Sensitivity and Adaptation* (*CS4, *SD3, *SD1, *CS1).

The previously mentioned output from GPT-4o includes two *Verbal Expressivity* items (VE3 & VE4) and two *Conversational Skill* items (CS2 & CS3) placed in a *new category* entitled by GPT-4o as "Engagement and Interest Management". Also, another new category named "Interpersonal Sensitivity and Adaptation" was created by GPT-4o which comprises two *Self-Disclosure* items (SD1 & SD3), as well as two Conversational Skill items (CS1 & CS4). After close inspection of their constituent items, these two new category labels generated by GPT-4o appear plausible and represent yet another *alternative interpretation* of the semantic content of a particular group of items and their association and generalization in constructs independent of the researcher's original theoretical assumptions.

The notion that the output from GPT-4o (as well as from other LLMs in our study) should be viewed as more *probabilistic* than *deterministic* was confirmed when the previous prompt was repeated with GPT-4o, but with the 16 items rearranged once more in a somewhat different order than that used in the first instance of item randomization discussed above. In response to this *third reproduction of the same basic prompt* but with a different item randomization (i.e. *another rearrangement of the sequence of items*) in the prompt, two new and somewhat different constructs appeared: "Social Interaction Management" (similar to but not the same as *Composure*) and "Conversational Harmony and Flow" (similar to but not the same as *Conversation Skill*).

To investigate the possibility of *confirmatory semantic analysis* of the questionnaire that measures the four interpersonal skills, the following prompt that used the original scale labels for the categorization was given to all the LLMs in our study: "Categorize 16 items related to the following four categories of interpersonal communication skills: (1) Verbal Expressivity, (2) Self-Disclosure, (3) Composure, (4) Conversational Skill. Place exactly four items in each category by choosing the semantically most similar items that best describe the skill that labels each category. Here are the 16 items: <continued with a randomized list of 16 items>". In almost all instances the four items per scale were placed in their respective category even with several differently randomized sequences of items. In fact, all the LLMs that were used in our study (GPT-4o, GPT-4, Copilot, GPT-3.5, Gemini 1.5 Pro, and Claude 3.5 Sonnet) perfectly categorized all 16 items into the four predetermined categories when the latter prompt was used. Besides that, GPT-4 provided a brief argumentation alongside each item for its listing in a specific category, and Gemini 1.5 Pro added a concise interpretation of the content for each item (in brackets). In contrast to the above task of distributing the given scale items into predefined categories (which can be viewed as a form of a *confirmatory semantic analysis* of questionnaire scales), the previously described requests to LLMs to categorize scale items into undefined categories (regarding the names/labels for categories, the number of categories, and the number of items per category) can be viewed as forms of *exploratory semantic analysis* of questionnaire items and scales.

Having in mind the previously described examples of the use of GPT-4o for the categorization of items of the four self-assessment scales, a *tentative positive response can be made regarding the second research question*: "RQ2: Can prompts be used with GPT-4o and other advanced LLMs to categorize the items of questionnaires with multiple self-assessment scales in a form that complements their empirical evaluation using confirmatory factor analysis?". Apparently, GPT-4o and other advanced LLMs can be of assistance in such categorization tasks in case of (1) a small number of scales (constructs/categories) and (2) a limited number of items per scale, but this has to be further tested with other types of scales and questionnaires. In any case, such semantic analyses and refinements using LLMs are not difficult to perform *before* or *after* self-assessment scales or questionnaires are applied to human subjects to explore their construct validity.

## 6.3 Use of LLMs for the improvement of self-assessment scales

To further test GPT-4o at a much larger scale, eight positively worded items from the selected 17 self-assessment scales of the ICCI were scrambled in a circular sequence (A1, B1, C1, D1, […], A2, B2, C2, D2, […], A3, B3, C3, D3, …) and placed in the following prompt (with slightly adapted labels of some scales to improve the definitions of the constructs) for GPT-4o: "Categorize the 136 items related to the following 17 categories of interpersonal communication skills: (1) Initiation of Interaction, (2) Verbal Decoding of Messages, (3) Nonverbal Sensitivity, (4) Adaptability in Communication, (5) Composure in Interaction, (6) Self-Disclosure, (7) Empathy, (8) Comforting and Support, (9) Verbal Messages Encoding Skill, (10) Nonverbal Expressivity, (11) Verbal Expressivity, (12) Interaction Involvement, (13) Self-Monitoring, (14) Conversational Skill, (15) Assertiveness, (16) Interpersonal Control, (17) Interaction Management. Place exactly 8 items in each category by choosing the semantically most similar items that best describe the skill that labels the category. Use each item only once. Here are the items: <a sequence of 136 items completes this prompt>". This was a somewhat ambiguous task due to certain overlap between the following sets of skills in actual communication behaviors that contribute to those skills: (a) Empathy & Comforting and Support; (b) Assertiveness & Interpersonal Control & Interaction Management; (c) Verbal Messages

Encoding Skill & Verbal Expressivity & Conversational Skill. Also, it can be assumed that, at least for some self-assessment scales, the theoretical classification of items in ICCI scales, at the time this research questionnaire was designed, had at least some deficiencies and imperfections.

The correspondence between the theoretical classification and the categorization of 136 items into 17 categories of skills performed by GPT-4o was as follows (in the brackets, the first number refers to the number of items classified by GPT-4o that equals the theoretical (8 positively worded items were in all scales listed in the prompt!), followed by the number of plausible/relevant classifications that were theoretically assigned to a different skill/category but whose classification can still be deemed relevant/plausible): Initiation of Interaction (7+0), Verbal Decoding of Messages (6+1), Nonverbal Sensitivity (6+0), Adaptability in Communication, Composure in Interaction (4+2), Self-Disclosure (8+0), Empathy (6+2), Comforting and Support (7+1), Verbal Messages Encoding Skill (4+2), Nonverbal Expressivity (5+1), Verbal Expressivity (2+0), Interaction Involvement (3+0), Self-Monitoring (2+3), Conversational Skill (2+2), Assertiveness (4+2), Interpersonal Control (1+3), Interaction Management (1+6). A brief analysis of the previously listed results indicates that, on average, 4 out of 8 items were classified by GPT-4o as was theoretically assumed in the original ICCI questionnaire, and, on average, another 1.4 items were classified in another category that at least partly matched the content of the item. It must be noted that this was "zero-shot prompting" and that (a) the 17 categories of interpersonal communication skills ('constructs') had not been previously defined/explained to GPT-4o, (b) no previous example had been given to GPT-4o on how to perform this task, and (c) the ICCI questionnaire was not a part of the training data for the GPT-4o model (since the ICCI questionnaire and its items had not been earlier published). Having all of the previous in mind, as well as the fact that (1) some of the constructs at least partly overlapped and (2) the ICCI had not been refined to the diagnostic level (having been developed only for a specific research purpose), it can be concluded that, even though GPT-4o did not fully meet the expectations, it still provided the researcher with information about the potential overlap of constructs and misplacement of some items in scales whose constructs they poorly measure. It is evident that the original ICCI scales and their alterations could be tested for semantic consistency by GPT-4o and other LLMs without the need for further empirical application of the original ICCI scales on new human subjects.

According to a statement made in the *APA Guidelines for Psychological Assessment and Evaluation* (APA, 2020, p. 5), which can be interpreted as a recommendation to psychologists: "Research and ongoing studies continue to expansively and specifically increase our understanding of human behavior, cognition, and affect. As a means of measurement and evaluation of these factors, existing psychological tests are revised, and newly developed tests are designed to improve accuracy and validity of testing." It must be noted that the focus of the study that is presented in this paper *was not on personality psychology tests*, but on the self-assessment scales and questionnaires for the measurement of *interpersonal communication skills*, while the main motive for the research was to analyze the possibilities to improve such measures with LLMs like GPT-4o.

*Content validity* is an important component of the quality of a self-assessment measure. It is associated with the degree to which the items of an instrument represent a specific *construct* or the *conceptualization of the variable* they are designed to measure (Groth-Marnat and Wright, 2016, p. 18-19). Content validity can depend on the theoretical framework selected by the authors of a measure. On the other hand, *face validity* refers to the degree to which a measure appears to reflect a specific construct by those who are not necessarily knowledgeable of the construct, like the subjects in a study responding to the items of a measure (Furr, 2011, p. 54-55). It is emphasized that researchers should not rely only on face validity and assume that content validity does not have to be evaluated if face validity appears satisfactory.

One of the means to examine content and face validity was outlined earlier in this section and well as in *section 6.2* of this paper. It is the categorization of items that may indicate that, possibly, the wording of an item and its semantic content does not correspond to a particular construct since the item was placed by GPT-4o or some other LLM to a 'wrong' category, i.e. one that is different from the original category devised by the researcher. This can also suggest that specific constructs and items for their measurement have a 'semantic overlap'. In that respect, the following type of prompt was used to examine content validity with GPT-4o, which in this specific example uses three ICCI scales with potential construct overlaps: "Assertiveness is <definition of skill>. Interpersonal Control is <definition of skill>. Interaction Management is <definition of skill>. Categorize the following 48 items that are related to (a) assertiveness, (b) interpersonal control, and (c) interaction management into these three categories depending on the items' semantic content. Place 8 positively worded and 8 negatively worded items in each of the three categories. Here is the list of 48 items to be categorized: <randomized list of items for the of the 3 self-assessment scales>".

After the previous form/model of wording with the *Assertiveness*, *Interpersonal Control*, and *Interaction Management* ICCI scales was used with GPT-4o, the analysis of its response to the prompt indicated to the author/researcher a likely semantic overlap of these three constructs. After the re-examination of the previously collected empirical data (N=403; 72% male and 28% female; first- and second-year students of information systems; data collected in the year 1999) wherein only the data for these three scales and their positively worded items were subjected to factor analysis, the author of the ICCI measure concluded that most of their positively worded items could be grouped under a new joint construct labeled 'Interpersonal Influence and Interaction Management'(see *Appendix II* for the items that could constitute that adapted self-assessment scale for the measurement of this 'joint construct'). In case of a potential future redesign of the ICCI instrument, a new joint construct with such a label could be one of the choices to explore. The brief example above illustrates one of the methods the researchers can explore for the improvement of *content validity* of self-assessment scales with the use of feedback from GPT-4o and other advanced LLMs regarding the semantic content of their items and self-assessment scales.

Finally, slightly modified prompts were given to GPT-4o to test the possibility of a different semantic interpretation of the associations among the items of *Assertiveness*, *Interpersonal Control*, and *Interaction Management* scales: "Categorize the following 24 positively <negatively> worded items related to communication skills into only two categories and name each of the categories with the best possible label. Place any number of items in each category but do not use the same item twice (place it in both categories). Also, use for categorization only those items that fit a category in the best way, that is, some items may remain uncategorized. Here are the 24 items: <a list of 24 items>". This kind of re-examination of the items of the three ICCI scales *(Assertiveness, Interpersonal Control, and Interaction Management)*, performed separately for positively and negatively worded items, resulted in plausible categorizations. For the *positively worded items*, the category labels provided in the response from GPT-4o were: (1) Influence and Persuasion and (2) Effective Communication and Relationship Management. For the *negatively worded items* the categories were similar (but in *reverse meaning* from the labels above): (1) Ineffective Influence and Assertiveness and (2) Poor Communication and Relationship Management. These responses seem to correspond with the previously mentioned results of the re-examination of empirical data using factor analyses of the positively worded items of the three original ICCI scales (*Assertiveness, Interpersonal Control*, and *Interaction Management*) in which a potential new joint construct labeled "Interpersonal Influence and Interaction Management" emerged (see *Appendix II*).

When examining the answers to the prompts obtained from GPT-4o, it must be noted that they should be used only for *heuristic purposes*, as alternative perspectives or interpretations that have to be tested *cum grano salis*, i.e. with a critical stance. In fact, in these cases, the responses by LLMs to the prompts created a form of a problem-solving situation for the researcher who has to answer the following question: *How much can I trust and rely on the output from an LLM and what should I do with that output?* To illustrate this problem, two prompts were given to GPT-4o in a new session, with two lists comprising the same items with positive and negative wording, respectively (the 24 positively worded and 24 negatively worded items were identical to those ICCI items of the *Assertiveness, Interpersonal Control*, and *Interaction Management* scales in the prompts discussed above). The first prompt, which included only positive items was: "Categorize the following 24 positively <negatively> worded items related to unidentified flying objects (UFOs) into only two categories labeled 'Hostile UFOs' and 'Friendly UFOs'. Place any number of items in each category but do not use the same item twice (place it in both categories). Also, use for categorization only those items that fit a category in the best way, that is, some items may remain uncategorized. Here are the 24 items: <list of positively/negatively worded items>". Not surprisingly, in the responses to two prompts obtained by *GPT-4o*, the lists of the *uncategorized items* had only 4 positively and 12 negatively worded items, even though none of the 24+24 items were related to UFOs. With *GPT-4* the output was similar but with the following remark: "I will focus on the connotations behind each statement to determine whether they suggest a more hostile or friendly approach in interactions". *Copilot* performed the task and added several UFO-explaining web links. *Claude 3.5 Sonnet* categorized only 14+14 out of 24+24 positively and negatively worded items as related to UFOs but without any comments. Finally, *Gemini 1.5 Pro* was the only LLM that demonstrated some degree of 'critical thinking' since it categorized most of the positively worded and 50% of the negatively worded items as UFO-related but with the following remarks: "…categorizing these negatively framed statements into 'Hostile UFOs' and 'Friendly UFOs' is a stretch. These statements describe social awkwardness, lack of assertiveness, and difficulty with manipulation – traits that don't easily map onto the concept of extraterrestrial beings, whether hostile or friendly."; "However, we can attempt a loose categorization based on how these negative traits 'might' be perceived in the context of UFO interactions, with the HUGE caveat that this is entirely speculative and based on extremely limited information…"; "Attributing human-like qualities and motivations to hypothetical alien beings is highly speculative and prone to anthropomorphism (projecting human traits onto non-human entities). These categorizations are for a thought experiment only and shouldn't be taken as serious analysis of potential UFO behavior."

In popular terms, regarding the UFO-related output, GPT-4, GPT-4o, Copilot and Claude 3.5 Sonnet may be deemed to have 'hallucinated' or 'confabulated', probably due to their 'suggestibility' (i.e. susceptibility to inaccuracies in user-generated prompts), resulting in a 'confident' but incorrect output resulting from their intention to fulfill the prompt (see: Berberette et al., 2024). Nevertheless, there are arguments that even confabulations of LLMs may have a potential value (Sui et al., 2024). For instance, even though it is not commonly thought about in UFO-related popular discourse, with some thinking 'out-of-the-box' regarding the output from the LLMs, UFOs could be imagined as potentially 'hostile' or 'friendly' probes with AI 'programmed' to communicate using some form of communication skills with the inhabitants of explored planets.

The previous examples indicate potential problems with the use of LLMs for the improvement of self-assessment scales for the measurement of interpersonal communication skills. Despite that, since it was also demonstrated that their use can have a potentially positive outcome (i.e. the reconstruction of the *Assertiveness, Interpersonal Control,* and *Interaction Management* scales of the ICCI), it may be opportune to examine various additional approaches to self-assessment scale improvement as well.

One additional method of examining potential improvements to self-assessment scales using LLMs would be to ask GPT-4o (or other LLMs) to identify one or more items that are *least semantically related to other items in the scale and the construct measured by the scale*. As an example, such a prompt may be formulated as follows: "The following items are from four interpersonal communication skills self-assessment scales that are named: 'Verbal Expressivity', 'Self-Disclosure', 'Composure', 'Conversational Skill'. There are four items in each scale. Choose one item in each scale that is semantically least related to other items in the scale (and least related to the meaning of the scale label) and explain why the chosen item doesn't belong with the other items in a specified scale. Repeat this for all four scales. The scales and their constituent items follow. Verbal Expressivity (VE) scale: VE1 – I can easily engage others by recounting various interesting events and anecdotes. <in the continuation is the list of other items and scale labels>".

In its response to the prompt above, GPT-4o identified one item from each scale and provided an explanation for such choices. For instance, similarly to the data exhibited in *Table 2*, it correctly identified that for the *Self-Disclosure* scale the following item is the least suitable representation: "SD2 – By selecting personal thoughts and feelings that I share with others, I appear more attractive to them and gain their favor." Here is the explanation that GPT-4o gave for its response in this specific case: "SD2 emphasizes using self-disclosure as a means to gain favor and appear more attractive, which introduces a manipulative or strategic element. The other items (SD1, SD3, SD4) focus more on the appropriate and sensitive sharing of personal information to build closeness or maintain appropriate boundaries, aligning more with the concept of self-disclosure in interpersonal communication." The same kind of seemingly plausible explanation was given for the three items that least represent each of the three other scales.

LLMs like GPT-4o can also be asked to further elaborate on their response to the prompt given above, enabling a sort of dialogue about specific items in self-assessment scales. The following prompt was subsequently given to GPT-4o: "In the Self-Disclosure (SD) scale you identified 'SD2 – By selecting personal thoughts and feelings that I share with others, I appear more attractive to them and gain their favor.' Which other scale would it be best categorized to?" Without reflecting on its theoretical correctness, the following response of GPT-4o demonstrated a considerable resemblance to the 'theory of mind' level of reasoning ability (see: Strachan et al., 2024): "The item 'SD2 – By selecting personal thoughts and feelings that I share with others, I appear more attractive to them and gain their favor.' would be best categorized under the 'Conversational Skill (CS)' scale. Here's why: The Conversational Skill (CS) scale focuses on the adeptness in managing conversations to achieve certain interpersonal goals and harmonizing interactions. SD2 involves strategically selecting what personal information to share to create a favorable impression and gain social approval, which aligns well with the skills required to manage and navigate conversations effectively. In particular, it aligns closely with the goals of: 'CS2 - I can redirect the topics of conversation with others so that everyone achieves their needs and goals.' This item emphasizes the ability to steer conversations to fulfill mutual goals, akin to selecting personal disclosures to achieve the goal of appearing more attractive. Overall, SD2's focus on strategic self-disclosure for social gain fits well with the objectives and skills encapsulated in the Conversational Skill scale." The previous example of a response generated by GPT-4o illustrates the potential to use LLMs in the form of a dialogue about self-assessment scale items and related constructs.

It must be noted that *GPT-4*, *Copilot*, *Claude 3.5 Sonnet*, and *Gemini 1.5 Pro* responded similarly well as *GPT-4o* to the previous two prompts, also identifying the item SD2 as least similar to the "Self-disclosure" scale, as well as suggesting its categorization in the "Conversational Skill" scale. However, the wording and arguments provided by all LLMs were somewhat different.

To improve self-assessment scales the following type of prompt could also be used to extend the dialogue with those LLMs: "Suggest one original new item for each scale (with a different wording and semantic content) but in a way that it still aligns with other items in terms of what they measure and the general construct/label of the scale. Explain your choice for every new item that you propose". After receiving the response to such a prompt, the scale author/researcher could then choose among the best proposed new items by the LLMs. Eventually, in the continuation of the dialogue, the following prompts could be used after that given above: "Try to make the new item<s> as different as possible from other items in the scale, while still being a component of the scale."; "Now slightly simplify the latest version<s> of the item<s>." Of course, various other prompts can also be delivered to LLMs that can support the improvement of various aspects of self-assessment scales and their items.

Regarding the previously demonstrated comparable respective capacity of GPT-4o, GPT-4, Copilot (GPT-4 Turbo), Gemini 1.5 Pro, and Claude 3.5 Sonnet to classify, interpret, rephrase, and semantically analyze the items of interpersonal communication skills self-assessment scales, as well as to create new items regarding the related constructs, *a positive response can be given to the third research question*: "RQ3: Can GPT-4o and other advanced LLMs be effectively utilized in activities like evaluation, improvement, and redesign of interpersonal communication skills self-assessment scales and questionnaires?"

## 6.4 Design of new self-assessment scales using LLMs

As was mentioned earlier, in the field of personality psychology the possibility to generate Big Five personality items for self-assessment scales was demonstrated using *GPT-3* (Lee et al., 2023) with a selection of final items by human evaluators, as well as with the use of more advanced *ChatGPT* to create a 'ready to use' short version of the Big Five measure with a single prompt (Walton and Anguiano-Carrasco, 2024) without human expert intervention in the measure generated by ChatGPT. It must be noted that Franco-Martínez et al. (2023) illustrated how prompts for LLMs can be used for generating items for the measurement of non-cognitive constructs. Correspondingly, it can be assumed that advanced LLMs like GPT-4/GPT-4o, Copilot (GPT-4 Turbo), Gemini 1.5 Pro, and Claude 3.5 Sonnet can also be utilized for generating items for interpersonal communication skills self-assessment scales.

For instance, testing of the capacity of LLMs to *generate a new short-form personality self-assessment measure that is related to interpersonal communication*, without any preparation of the LLM (e.g. with a zero-shot prompt), but with the use of the International Personality Item Pool (IPIP) for model items (IPIP, 2023) and with a specific choice of personality trait labels (a researcher can choose those from: IPIP, n.d.) *can be addressed within a single prompt*: "By using the International Personality Item Pool create five totally new representative items for each of the following five interpersonal communication skills related self-assessment scales from the alphabetical index of 274 labels for IPIP scales: (1) Compassion, (2) Fairness, (3) Politeness, (4) Impression Management, (5) Enthusiasm. Make the content of the items applicable for self-assessment in the context of face-to-face interaction with other people. The five-point Likert scale for the responses is in the range from '1 = Not at all true of me' to '5 = Very true of me'." Regarding the apparent *face validity* of the produced items and their respective scales, the five examined LLMs (of GPT-4o, GPT-4, Copilot, Gemini 1.5 Pro, and Claude 3.5 Sonnet) performed fairly well at this task, as judged by the author who is a psychologist and also an expert in interpersonal communication skills. For brevity purposes, here are only sample items that the LLMs produced after the use of the above prompt for the creation of a *Compassion* scale:

- *GPT-4o:* "I often notice when someone is feeling down and try to comfort them."
- *GPT-4:* "When I see someone upset, I instinctively try to make them feel better."
- *Copilot:* "I genuinely care about others' well-being."
- *Gemini 1.5 Pro:* "When someone shares their problems with me, I can easily put myself in their shoes and feel their emotions."
- *Claude 3.5 Sonnet:* "I often find myself offering comfort to others during difficult conversations."

It would be an interesting experiment to ask the aforementioned LLMs to perform a similar scale-generating task with some of the other 200+ labels for IPIP scales (IPIP, n.d.) and subsequently have human experts evaluate the results. However, for further analyses in this paper, a similar prompt was worded regarding some of the scales of the ICCI that had not been earlier published and are also *not related* to constructs previously mentioned in the earlier sections of this paper or used in dialogues with the LLMs by the author: "For the following constructs related to interpersonal communication skills write a brief definition of each construct and then, based on your definition and other available scholarly resources and knowledge about these skills create five totally new representative items for each: (1) Initiation of Interaction, (2) Adaptability in Communication, (3) Interaction Involvement, (4) Verbal Decoding of Messages, (5) Nonverbal Sensitivity. Make the content of the items applicable for self-assessment in the context of face-to-face interaction with other people. The five-point Likert scale for the responses is in the range from '1 = Not at all true of me' to '5 = Very true of me.' Not surprisingly, having in mind the earlier demonstration with the selected IPIP items and scales, all LLMs again responded fairly well to this *zero-shot* prompt. For illustration, the best *construct definitions*, as well as *selected generated items* (chosen by the author) for each scale and particular different LLMs are listed here:

*Initiation of Interaction* (*GPT-4*): "Initiation of interaction refers to the ability to start a conversation or interaction with others. It involves taking the first step to engage someone in communication, which can include introducing oneself, asking a question, or making a comment to begin an exchange." // Item: "I find it easy to come up with topics to talk about when meeting someone for the first time."

*Adaptability in Communication* (*Copilot*): "Adaptability in communication refers to the ability to adjust one's communication style, tone, and content based on the context, audience, and situation. It involves being flexible and responsive in how we express ourselves." // Item: "I can easily switch between formal and informal language depending on the setting."

*Interaction Involvement* (*Claude 3.5 Sonnet*): "The degree of engagement and attentiveness one demonstrates during interpersonal exchanges, including active listening and responsive participation."; Item: I often ask follow-up questions to deepen my understanding of what others are saying."

*Verbal Decoding of Messages* (*Gemini 1.5 Pro*): "The ability to accurately interpret and comprehend the meaning behind the spoken words of others." // Item: "I pay attention to the context of a conversation to better understand the meaning behind the words."

*Nonverbal Sensitivity* (*GPT-4o*): "Nonverbal Sensitivity refers to the ability to perceive and interpret the nonverbal cues of others, such as body language, facial expressions, gestures, and tone of voice. It involves recognizing and understanding the emotions and intentions behind these nonverbal signals." // Item: I am good at reading other people's body language to understand how they feel."

It must be emphasized that all the brief definitions of each of the five specific constructs that were provided by respective LLMs were differently worded, while still resembling in their meaning to those that can be found in scholarly literature. Also, none of the definitions of any of the constructs was incorrect. Most of the items for specific scales of the five previously mentioned construct measures that were generated by various LLMs were different in wording as well. A large majority of the *generated items* by all the LLMs and for all the five scales semantically corresponded to the respective *theoretical definitions of the constructs*. It must be remarked that traditional scale development takes substantial research time to collect relevant literature, analyze the selected constructs, and create an initial pool of items related to selected interpersonal communication skills. With the use of the preceding prompt and the five examined LLMs, 25 items (5 for each of the scales mentioned in the prompt) were generated in about one minute by each LLM.

In addition to the previously revealed time-saving convenience for researchers who are developing interpersonal communication skills scales, it is also possible to ask LLMs to evaluate early/preliminary versions of such scales and item pools. Furthermore, the output from one LLM can be evaluated by another, for instance, the items created by GPT-4o could be analyzed by Gemini 3.5 Pro and vice versa (as well as in other possible combinations of LLMs). Here is the potential wording of a related prompt (for the list of scales and their constituent items the output of some other LLM is used): "For the following five interpersonal communication skills analyze the items in representative scales regarding their potential to measure the associated skill as well as how well each item semantically relates to other items in the same scale: (1) Initiation of Interaction, (2) Adaptability in Communication, (3) Interaction Involvement, (4) Verbal Decoding of Messages, (5) Nonverbal Sensitivity. The content of the items should be related to self-assessment in the context of face-to-face interaction with other people and in concordance with the respective constructs that can be found in scholarly literature. The five-point Likert scale for the responses would be in the range from '1 = Not at all true of me' to '5 = Very true of me.' Here are the scales and their items <a list of scale labels followed by their constituent items generated by some other LLM>." After testing this prompt, the researcher concluded that the best-performing LLMs in this study, *GPT-4o*, and *Gemini 3.5 Pro*, provided informative and extensive feedback to the researcher about the scales in general, as well as of the items that were generated by the other LLM. The same was also achieved using *Claude 3.5 Sonnet* after the previous prompt was supplemented with the following remark after the first sentence: "**Do this for the scales and each of their items separately.**"

Finally, in a more controlled approach with multiple-shot prompting for a single new scale (i.e. with more than one example delivered in a prompt before generating a new scale with an LLM, as well as with greater previous engagement of the LLM regarding the items and the corresponding construct) the prompts could consist of the following:

1) Definition of a single construct for which new measurement items should be generated;

2) Extensive list of exemplary items;

3) Demand for the evaluation of each of the exemplary items regarding their association with the definition of a specific construct and other criteria like semantic alignment with other exemplary items for the specific construct;

4) Finally, request for the creation of new items for the measurement of the construct.

As an illustration of the above, here is the first-stage prompt that was tested with five LLMs including the 25 items from the unpublished *Active Listening* scale designed by the author (the text of the extensive 83-word definition of the *active listening* construct was copied from Wikipedia; the items of the *Active Listening* scale and the responses to this prompt by each of the LLMs is displayed in *Appendix III*): "Active listening is <extensive definition of the construct from the Wikipedia>. Act as an expert regarding the interpersonal communication skill 'active listening'. Your assignment is to judge 25 items that may be related to this concept/construct. For each item give one of the following evaluations: 1 = Irrelevant {for active listening}; 2 = Important, but not essential {for active listening}; 3 = Essential {for active listening}. Write your evaluation on the left side of each item. The 25 items to evaluate are: <following is a list of 25 original items>". After the LLMs provided their responses for each item (see *Appendix III* for the numerical responses), in the next/second stage they were given the following related prompt to generate a completely new 10-item active listening scale: "Having in mind the definition of active listening and the previous 25 items as examples of active listening, create a totally new active listening scale with only 10 best possible items and without repeating the content of any of the 25 items that were used as an example. After each new item briefly explain the choice for the wording of that item." Here are the samples of some of the best new active listening items generated by each of the LLMs that were used in our study:

- **GPT-4o:** "I adjust my listening style based on the speaker's needs, such as being more patient with those who struggle to express themselves".
- **GPT-4:** "I summarize what the speaker has said to ensure my understanding is correct".
- **Copilot:** [Struggled with the second prompt]
- **Gemini 1.5 Pro:** "I make a conscious effort to understand things from the speaker's perspective, even if I don't share their views".
- **Claude 3.5 Sonnet:** "I provide verbal acknowledgments (like 'I see' or 'mm-hmm') to show I'm engaged without interrupting".

The example above illustrates an approach in which the LLMs are provided with more information through sample items and a definition of the construct for which they are required to generate new items, i.e. a *new scale* or *novel items* for a scale. The categorization of items during their initial evaluation as "Irrelevant", "Important, but not essential" and "Essential" was made according to Lawshe (1975), but the categories can be labeled differently. It must be noted that the request in the first-stage prompt presented earlier in this section can also be used to experiment with Q-sorting of scale items using different LLMs as separate judges/experts. Although this procedure of using the LLMs in the capacity of experts/ judges was apparently successful regarding the 'confirmatory semantic analysis' of the scales and items for the measurement of the four communication skills (*Verbal Expressivity, Self-disclosure, Composure, Conversational Skill*) in *section 6.2* of this paper, in the case of evaluation of the 25 items of the *Active Listening* scale considerable disagreements were found between the five LLMs (*GPT-4o, GPT-4, Copilot, Claude 3.5 Sonnet*, and *Gemini 1.5 Pro*) that can be observed from the table shown in *Appendix III*.

Regarding the previously presented examples of item generation for various interpersonal communication skills constructs, it is possible to *affirmatively respond to the fourth research question of the study that is addressed in this paper*: "RQ4: Can GPT-4o and other advanced LLMs generate/design new interpersonal communication skills self-assessment scales and questionnaires?"

## 6.5 Automated application and scoring of self-assessment scales with LLMs

LLMs like *GPT-4o, GPT-4, Copilot, Claude 3.5 Sonnet*, and *Gemini 1.5 Pro* are essentially very advanced forms of chatbots or *multi-turn dialogue systems* (see: Yi et al., 2024). Some benchmarking studies have found GPT-3.5 and GPT-4 to outperform other LLMs in *multiturn tasks* (see: Kwan et al., 2024; Chen et al., 2023), but they did not include evaluations of Gemini 1.5 Pro and Claude 3.5 Sonnet. In this section, the potential to *use a single prompt* to instruct LLMs to perform the *design and application of new interpersonal skills scales* (or questionnaires) will be briefly demonstrated. It must be emphasized that, for this purpose, it is opportune to use at least one-shot or, preferably, multiple-shot prompting to ensure a better-quality new self-assessment instrument. Also, at the current level of LLM development and without proper ethical guidance it is recommended to employ an LLM for the automated application of interpersonal skills scales only for exploring the capacities of a particular LLM while avoiding any diagnostic use or interpretations except for research purposes.

Here is a general form of a potential zero-shot prompt that worked well with almost all the LLMs that were examined (i.e. *GPT-4o, GPT-4, Gemini 1.5 Pro, Claude 3.5 Sonnet*): "Using the best possible items for self-assessment of the communication skill 'Active Listening' create a brief new scale with 10 items regarding the use of active listening in everyday communication. Use a 1 to 5 Likert scale ranging from '1 = Not at all true of me' to '5 = Very true of me' for collecting responses. After you have created this new scale, ask me one by one questions and wait for my answers from 1 to 5. When you receive my responses to all 10 questions, summarize my score and provide me with feedback about my active listening communication skill." After this prompt, each of the examined LLMs created a considerably different set of 10 items related to *active listening* with apparent *face validity* and then applied the items one by one (by asking each question separately and waiting for the response). After collecting all 10 responses in a dialogue with the user, the LLMs summarized numerical responses and interpreted the 'total score' of the user. The same form of prompting can work for various interpersonal communication skills and other types of related self-assessment topics. However, it must be mentioned that *Copilot* initially performed this type of task very well for weeks. Still, around the end of July 2024, it suddenly lost the ability to adequately respond to the aforementioned prompt in the form of a one-by-one multiturn application of all 10 created items (although this inability could have been a temporary issue).

Initial versions of items and self-assessment scales can be produced and applied by the LLMs that were used in this study for various other topics beyond interpersonal communication skills. However, researchers must keep in mind that the related constructs should be relevant and the psychometric quality of the new measure needs to be empirically tested. For instance, the following prompt, similar to the preceding, initiates the 'design' of a mock ('made-up') self-assessment scale with an immediate online application of its items: "Create a self-assessment scale related to the ability to theatrically recite poetry in front of an audience. The responses to the items should be on a Likert scale from 1 to 5, where 1 is 'Not at all true of me' and 5 is 'Very true of me.' Ask me the 10 questions in a sequence, one by one question, requesting my response on the Likert scale, and finally summarize my responses with a brief interpretation." Some of the LLMs may perform better at such tasks and, at the time this study was conducted, the best results, according to the experiences of the author, were yielded by the use of GPT-4o, Gemini 1.5 Pro, and Claude 3.5 Sonnet.

Finally, here is a general form of a prompt to request the design of a questionnaire with more than one scale, as well as its immediate application, both of which should be experimented with the most capable LLMs available (like GPT-4o, Gemini 1.5 Pro, and Claude 3.5 Sonnet at the time of this study, or newer and more advanced): "Create self-assessment scales with <N> items each that are related to the following interpersonal communication skills: (1) <construct name for first skill>, (2) <construct name for second skill>, (3) <construct name for third skill> <etc.>. The responses to the items should be on a Likert scale from 1 to 5, where 1 is 'Not at all true of me' and 5 is 'Very true of me.' <or some other response scale>. Make sure that you ask me all of the <N x number of constructs> questions in a sequence, one question at a time, reminding me to respond using the Likert scale with each question. After collecting all my responses to all of the items, summarize them separately for each skill with a brief interpretation."

Having in mind the period in which the LLMs in our study were examined (June - August 2024), concerning the generation of new scales for the assessment of interpersonal communication skills and their automated application it is possible to provide a *conditionally positive response to the fifth research question*: "RQ5: Can GPT-4o, Gemini 1.5 Pro, and Claude 3.5 Sonnet generate new scales and respective items for the measurement of interpersonal communication skills, and subsequently apply them to users in an interactive multiturn form?" The response to RQ5 is *tentatively positive* since it depends on the type and availability of the LLMs used as resources, and probably also on the construct for which a scale is designed, the number of scales in a multifaceted questionnaire, as well as the number of items in a specific scale. For instance, there could be a limitation regarding the number of prompts that can be delivered to an LLM which restricts the number of items that can be successively delivered to the user in an automated manner. Also, it must be noted that regarding some topics and potentially 'controversial' constructs (like the construct 'an obsession with conspiracy theories', for example) an LLM may decline the request to design and interactively deliver an assessment scale in an automated form.

## 6.6 Practical advice regarding the use of LLMs for translation, evaluation, improvement, and design of new self-assessment scales

Effective use of LLMs commonly requires some practice and experience. Expert advice is also beneficial, as well as learning from *exemplary prompts* that can be found in the instructions of the providers of LLMs for their use, on the web, and in other resources (e.g. templates, guides, and tutorials for *prompt engineering*). In various sections of this paper, a collection of useful prompts is provided for specific objectives which, however, only represent a sample of a wide variety of potential prompts that can also serve a similar purpose.

Various LLMs may demonstrate an advantage in different particular tasks. For instance, with its responses to the prompts *Copilot* often provided *potentially useful links* to related web pages. Also, *GPT-4o*, *Gemini 1.5 Pro*, and *Claude 3.5 Sonnet* could perform more sophisticated dialogues about interpersonal communication skills and constructs compared to less advanced LLMs. Easy access to archives of previous dialogues with an LLM can also be beneficial for different purposes, as well as the possibility to share a link (access) to completed and archived previous dialogues of the researcher with an LLM. It is often opportune to explore using the same prompts with various LLMs since their answers may vary and usually do not have a complete semantic overlap in the content provided. For instance, the goal to create a *most diverse pool of initial scale items* for the measurement of a specific construct with the aid of LLMs will probably be better achieved by using multiple LLMs.

At the time the practical part of this study was conducted the five examined LLMs (GPT-4o, GPT-4, Copilot, Gemini 1.5 Pro, and Claude 3.5 Sonnet) were easily accessible and well-functioning. Also, their performance was similar according to benchmarking tests (Hart, 2024; for an up-to-date LLMs leaderboard see: Chatboard Arena, 2024), including multilingual evaluation (Wiik, 2024). The similar level of functioning of examined LLMs during the time this study was conducted was confirmed by the fact that they were able to compete with each other in simple games (Topsakal et al., 2024). The comparable level of sophistication enabled *the parallel use of LLMs for generating and evaluating self-assessment scales and their items*, as well as the use of one explored LLM to evaluate the output of another, perform back-translation, etc. However, it is possible that 4-5 almost equal LLM market 'contestants' that can be utilized for various scientific analyses and scale item generation may not be equally available to researchers in the future as was the case at the time this study was completed in August 2024.

Several significant changes in the LLM arena were evident in the brief period from May to August 2024: (a) Copilot's conversational options "Balanced", "More creative" and "More precise" were canceled; (b) GPT-3.5 ceased to be publicly available; (c) GPT-4o and GPT-4o Mini were introduced; (d) Antrophic's Claude 3.5 Sonnet was introduced; (e) various new models were introduced by mayor IT companies, e.g. Meta AI's Llama 3 and xAI's Grok, some of them not widely available in all countries, etc.

Numerous previous studies related to the 'cognitive capacities' of LLMs, personality psychology, and related fields that were conducted with the use of less capable LLMs than GPT-4o, Gemini 1.5 Pro, and Claude 3.5 Sonnet could have become at least partly outdated in a matter of months, which also likely refers at least to some parts of the study that is presented in this paper. Despite the hype associated with rapidly advancing LLMs, researchers may consider addressing research topics that would, in their results, be longer-lasting than a half- to one-year phase in the evolution of LLM systems. For instance, regarding the study that is presented in this paper, it is already common knowledge that advanced LLMs perform fairly well in machine translation tasks and, probably, in the near future, if it is not the case already, their potential use for item generation in assessment scale development will become a regular practice.

Considering the evaluation scales and items of interpersonal communication skills self-assessment measures with LLMs, it is recommended that researchers use (a) their measures, (b) measures that are available for research without any restrictions, or (c) measures for which they have attained approval by their authors or copyright owners. To experiment with LLMs the International Personality Item Pool (IPIP, 2023) can be freely used, according to the statement on the IPIP website: "Because the IPIP has been placed in the public domain, permission has already been automatically granted for any person to use IPIP items, scales, and inventories for any purpose, commercial or non-commercial".

For the best choice of an LLM regarding the performance in *language translation* of items of self-assessment scales, it is recommended to consult related up-to-date benchmarks and tests that can be found on the internet. Also, multiple LLMs can be asked to perform the same translation tasks, after which the most satisfactory translation of each item can be selected. Some LLMs may perform a more *literal* translation that may better preserve the meaning of the original item but would not appear *natural* in the target language. In the context of scale/item development, the goal of translation could be conceived as twofold: (a) producing a translation in the spirit of the *source* language and keeping the essence of the original meaning of the item, while (b) achieving a natural-sounding translation in the *target* language. Keeping the 'essence of the meaning' of an item that is written in the source language is perhaps best evaluated by a *trained human scholar with domain expertise in the given construct*, who is also a native speaker of the *source* language and has a good knowledge of the *target* language (or vice versa). Achieving a 'natural-sounding' translation in a target language is perhaps best evaluated by a *professional translator and language expert* who is also a *native speaker* of the domain language. In any case, evaluation of *forward* and *backward* translations can also be asked from the LLMs which typically did not perform the original translation. However, regardless of employing LLMs for the improvement of item translation, before using a scale or a questionnaire for research on human subjects and/or publishing the self-assessment measure, it is recommended to have a proper human expert review the final version of the translated measurement instrument.

To construct/design new items for self-assessment scales that measure interpersonal communication skills it is advisable to previously consult relevant scholarly literature, especially regarding specific skills, as well as publications with a more general approach like: Spitzberg (2003), Spitzberg and Cupach (2011), Graham and Titsworth (2010), Shue and Stamp (2019), Hargie (2019) and other. After gaining substantial theoretical knowledge of a specific skill or set of skills, including that of best evaluated and currently available measures, a scholar may not even need the LLMs to design new self-assessment scales and their items. However, as was demonstrated in this study, LLMs can definitely assist in that task.

For research purposes, a carefully selected *methodology for the development of self-assessment scales* to be assisted by LLMs should first be considered (this procedure is not psychometric 'quantum computing' since it is explained in most psychometry textbooks; for a concise and simplified outline see: Boateng et al., 2018; for detailed professional recommendation see: American Educational Research Association, American Psychological Association, & National Council on Measurement in Education, 2014). In case of *diagnostic us*e of self-assessment instruments or their utilization for applied non-research purposes, the *APA Guidelines for Psychological Assessment and Evaluation* (APA, 2020) should be implemented as much as possible. The *automated (i.e. computer-based) application* of self-assessment scales on human subjects with the use of an LLM is a questionable activity (even if it is only intended for designing online learning activities for students), so it would be advisable to consult sources such as the *Guidelines for Technology-Based Assessment* (International Test Commission, 2022). In addition, instead of using an LLM for the automated application of a self-assessment instrument, an offline *Python* computer program or *HTML* program for an online application can be created with the assistance of LLMs like GPT-4o.

# 7 Summary and discussion

The main goal of the study that is presented in this paper was to explore and provide examples of several procedures by which LLMs like *GPT-4o, GPT-4, Copilot, Gemini 1.5 Pro*, and *Claude 3.5 Sonnet* can be used for the improvement, evaluation, and design/construction of self-assessment scales in the scientific field of interpersonal communication skills.

In *section 1* of this paper named **"Introduction"** a brief overview of the literature is provided regarding the potential of using ChatGPT and other more advanced LLMs in various phases and fields of academic research (Chukwuere, 2024), including specific areas like *STEM motivation* (Donmez et al., 2023), *adoption of AI in higher education* (Rahman et al., 2023), and *teacher technostress* (Khlaif et al., 2023). Also, there has been a rapid increase in the use of LLMs for *psychology applications* (Ke et al., 2024). In this paper the LLMs have proved to be beneficial in the specific scientific field of *design and improvement of self-assessment scales for the measurement of interpersonal communication skills*.

A very high capacity to process *textual* verbal content is necessary if LLMs are utilized to design and improve self-assessment scales. This 'cognitive' ability of LLMs was previously illustrated in *section 2* entitled **"Performance of LLMs in verbal and reasoning tasks"**. Indeed, LLMs have demonstrated performance that is similar to or above the average human level in *university-level exams and aptitude tests* (OpenAI, 2024). Their *emergent abilities in human-like reasoning* are rapidly advancing (Hagendorff et al., 2023), including their ability to perform *diverse problem-solving* (Orrù et al., 2023) and *analogical reasoning tasks* (Webb et al., 2023). There are numerous studies indicating that LLMs may surpass human subjects in specific *intelligence tests* and that they can also manifest 'creativity' regarding particular *divergent thinking tasks* (Guzik et al., 2023; Hubert et al., 2024). LLMs like GPT-4 can *solve novel and difficult tasks* close to human-level performance in various disciplines, including *psychology* (Bubeck et al. 2023), demonstrating the capability to 'reason' about the emotional state and intentions of others in complex situations, possibly at an advanced level of 'theory of mind'. It must be emphasized that the manifested verbal and reasoning abilities of LLMs imply their *potential to be used for the design and improvement of self-assessment scales and questionnaires not only for the measurement of personality traits but also for the assessment of interpersonal communication skills*.

In *section 3* named **"LLMs and personality research"**, a concise review is made of studies related to the *assessment of personality traits*, including prediction and/or classification regarding the *Big Five* personality dimensions or *Myers-Briggs Personality Type* traits that were based on *writings* of individuals (Guinn, 2023), as well as *recent tweets* of individuals (Murphy, 2024), *Facebook posts* (Peters and Matz, 2024), and *essays and tweets* (Ji et al. 2023). LLMs also manifested the *ability to classify emotions* (Belkhir et al., 2023), *produce verbal responses to emotional situations* (Welivita and Pu, 2024), *deliver empathetic communication* in interaction with real-world patients (Luo et al., 2024), as well as demonstrate a high level of *emotional awareness* (Elyoseph et al., 2023) and *understanding of emotions* (Wang et al., 2023). Finally, the ability to *generate Big Five personality items for self-assessment scales* was demonstrated with GPT-3 (Lee et al., 2023) and ChatGPT (Walton and Anguiano-Carrasco, 2024). Various *examples of prompts* for GPT-4 were also provided (Franco-Martínez et al., 2023) for *generating items that refer to non-cognitive constructs* (e.g. *personality*) *in different forms*, as well as for the analysis and improvement of item wording, content evaluation, construct representation, and evaluation of internal structure of scales. Therefore, before the study that is presented in this paper, *in the field of personality traits the first steps had already been made that can motivate researchers to use LLMs for the design and improvement of interpersonal communication skills self-assessment scales.*

The **instruments** that were used as main resources in this study were selected among the LLMs that were accessible in the period from June to August 2024 with a free account or paid subscription: GPT-4o, GPT-4, Copilot, Gemini 1.5 Pro, and Claude 3.5 Sonnet. The comparable performance characteristics of as many as five different publicly available LLMs (without the need for their pre-training), at the time this study was performed, made this an unprecedented research opportunity. Also, the other *instruments* or resources in this study were *previously unpublished self-assessment measures* developed by the author of this paper: (a) selected scales and items of the **Interpersonal Communication Competence Inventory (ICCI)**, (b) four shortened versions of ICCI scales for the measurement of **Verbal Expressivity**, **Self-Disclosure**, **Composure**, and **Conversational Skill**, as well as the (c) **Active Listening** scale. The advantage of using previously unpublished self-assessment scales is that their items could not have been included in the training data for the LLMs that were used in this study. Also, there were no potential ethical concerns regarding the (mis)use of scales developed by other authors (note: the *ICCI* was developed by the author of this study in 1999, and the *Active Listening* scale in 1995, both on the basis of construct descriptions in literature).

Regarding the **research questions,** it must be emphasized that only selected model case studies are presented in this paper for exemplary purpose. In this study no thorough research investigation was made into (a) *forward translation* and *back translation* capabilities of LLMs for multiple languages, (b) the potential to use LLMs for comparison of *semantic classification analyses* of scales with their confirmatory factor analyses with a large number of different self-assessment multifaceted questionnaires developed by different authors; (c) the capacity of the LLMs to *generate/develop new items for questionnaires with a large number of different constructs/scales and with diverse numbers of constituent items* that are related to interpersonal communication skills, and also (d) the quality and representativeness of generated scales and items based on the *evaluation performed by human experts and independent researchers*.

In *subsection 6.1* entitled **"Use of LLMs for the translation and simplification of self-assessment scale items"** the potential for translation of scales from Croatian to English is demonstrated. While there is evidence that GPT-4 is comparable to (Jiao et al., 2023) or can outperform commercial systems for machine translation (Kalyan, 2024), as well as that it may assist professional translators (Siu, 2023), researchers have also found that it falls behind in performance when compared with translators of medium- and senior-level expertise (Yan et al., 2024). Since each LLM offers a slightly different translation of individual items, based on the experience of the author it is suggested that *multiple LLMs are used to translate scale items and that the apparently best translations of individual items are considered for the final evaluation and editing by expert human translators*. The simplification of items was tested in this paper for *English language versions* of the self-assessment scales, wherein it was proven that (with the use of GPT-4o, for example) a significant shortening of items (regarding the number of words and characters) of at least 30% and more is possible for the 27 observed ICCI scales with a total of 374 items that were translated from Croatian to English by a human non-expert translator. In addition, it must be noted that the brevity and phrasing of the shortened items can be manipulated by the wording of the related prompts for LLMs.

The *subsection 6.2* of this paper labeled **"Use of LLMs for the categorization of items of self-assessment scales"** includes a demonstration of potential uses of LLMs as an alternative means for *construct validation* of multifaceted self-assessment questionnaires. In this case study the *shortened forms* of the following ICCI scales were used: *Verbal Expressivity*, *Self-Disclosure*, *Composure*, and *Conversational Skill*. The results indicate that if a categorization of items of multifaceted questionnaires is asked for in a prompt given to GPT-4o *without stating the labels for the constructs/scales*, the output is likely to appear as a sort of *exploratory semantic analysis* where GPT-4o groups the items according to their semantic content and labels the groups of items in a way that may differ from the theoretical constructs for which the measurement scales and items were originally designed. In addition to the procedures and prompts that are featured in the presented case study, both the *number of constructs* and the *required number of items per scale* may be differently defined in the prompts. On the other hand, when both the construct labels and the expected number of items per scale were defined in the prompts, as in the case study presented here, all the observed LLMs (GPT-4o, GPT-4, Copilot, Gemini 1.5 Pro, and Claude 3.5 Sonnet) were likely to provide equal categorizations of items as was theoretically assumed for the original scales in Croatian (and also comparable to the confirmatory factor analysis of the data collected with the original scales on Croatian subjects displayed in *Table 2*, N=170). This type of prompting of the LLMs with defined labels for scales and specification of the number of items per scale can be considered as *confirmatory semantic analysis*. However, in our case study only a brief demonstration was made. To examine such demonstration of using LLMs for the design, improvement and translation of scales/items, further research should be conducted with diverse multifaceted questionnaires, various types and number of measured constructs, as well as with different number of items in their corresponding scales.

Perhaps the most applicable examples are described in *subsection 6.3* named **"Use of LLMs for the improvement of self-assessment scales"**. This subsection starts with a re-examination of the scales and items of the ICCI multifaceted questionnaire. First, selected scales of the multifaceted ICCI and the GPT-4o LLM were used to illustrate potential difficulties in the case of using a large number of construct labels for categories (17) to classify a large number of items

(136) with zero-shot prompting. Then, the potential overlap of constructs for three related ICCI scales (*Assertiveness*, *Interpersonal Control*, and *Interaction Management*) was investigated in the form of an *exploratory semantic analysis* using GPT-4o to analyze 3x8 positively and 3x8 negatively worded items of the three aforementioned scales. The uncovered semantic overlap of these three ICCI scales motivated the *reexamination of previously collected empirical data with factor analyses*, with the conclusion that, from the three original scales, a single new joint scale can be designed labeled 'Interpersonal Influence and Interaction Management' that can replace the use of the three original scales with overlapping constructs (see *Appendix II*). Also, the potential 'hallucinations' or 'confabulations' were illustrated in related categorization tasks when LLMs were asked to classify the items of the three previously mentioned scales for measurement of communication skills into two categories 'Hostile UFOs' and 'Friendly UFOs' (or to leave them uncategorized). Four of the LLMs (GPT-4o, GPT-4, Copilot, and Claude 3.5 Sonnet) performed this task without any suspect and placed at least 50% or more of the communication skills items into the two UFO categories even though UFOs do not possess communication skills, and only one LLM (Gemini 1.5 Pro) made a critical 'remark' regarding being asked to perform this task. In the continuation of the same subsection, several potential methods for the improvement of items and scales were also illustrated: (a) prompts that ask LLMs to identify the least semantically related items to other items in a scale and to the construct measured by the scale; (b) request to the LLMs to further elaborate on their response to the previous type of prompt; (c) performing dialogues with LLMs about specific items in self-assessment scales and their various characteristics; (d) giving a prompt to elicit creation of one or more original new items for the particular scales with an explanation of the newly proposed item(s) by the LLMs. Such dialogues may also include asking comments from the LLMs about the clarity and wording of the items, and how close their wording is to the natural language expression.

A considerable domain 'knowledge' and an apparently high level of the 'theory of mind' of the LLMs concerning interpersonal communication skills is further illustrated in *subsection 6.4* entitled **"Design of new self-assessment scales using LLMs"**. With the use of *zero-shot* prompts, all of the observed LLMs (GPT-4o, GPT-4, Copilot, Gemini 1.5 Pro, and Claude 3.5 Sonnet) were able to provide acceptable definitions of the selected interpersonal communication skills constructs (*Initiation of Interaction, Adaptability in Communication, Interaction Involvement, Verbal Decoding of Messages, Nonverbal Sensitivity*) and create new items for corresponding measurement scales. The LLMs were also able to deliver comments on individual items of the new scales if asked to do this with a particular prompt, and some even performed this interpretation of new items without being specifically requested to do so. In addition, LLMs can be used to analyze items generated for specific scales by *other LLMs*, providing a valuable capacity to assist researchers in evaluating interpersonal communication skills self-assessment scales. Since LLMs are capable of *rephrasing sentences*, they can be used for *generating parallel versions of scales*, i.e. based on a sentence of one scale item they can generate one or more other items with a very similar meaning but using a different wording in the sentence. Finally, LLMs can *contextualize a self-assessment scale for a different environment*. For instance, an assessment scale that was created for the *face-to-face* interaction context can be contextualized to the context of online Facebook communication. This exemplary contextualization task was performed rather well with all of the LLMs that were examined in our study (GPT-4o, GPT-4, Copilot, Gemini 1.5 Pro, and Claude 3.5 Sonnet) using the positively worded items of the previously mentioned ICCI scales: *Initiation of Interaction, Adaptability in Communication, Interaction Involvement, Verbal Decoding of Messages, Nonverbal Sensitivity*. The prompt that was used for that purpose and the result of the contextualization (five original items from the ICCI scales and samples of the new items generated by all the five LLMs) are presented in *Appendix IV*. However, when using LLMs for the automated generation of 'new' items, the researchers should be aware that, without sample items and/or very concise and extensive definitions of constructs the LLMs may use their training data, which potentially consists of previously published scales authored by other researchers and similar copyrighted material.

Briefly said, in *subsection 6.5* named **"Automated application and scoring of self-assessment scales with LLMs"** it was demonstrated that by using a *single well-worded prompt* most of the examined LLMs (GPT-4o, GPT-4, Gemini 1.5 Pro, and Claude 3.5 Sonnet) can both design a single new self-assessment scale and apply its items one by one with the scoring of the user's responses and evaluation of the total score. More advanced LLMs like GPT-4o can perform this task with multifaceted questionnaires (i.e. ones containing more than one scale and more than 5-10 items per scale, but these capabilities were not thoroughly tested with other LLMs in this study). It is advised that for designing new scales psychometric guidelines are followed. Also, for proper application of the newly designed scales for a diagnostic purpose with humans demanding criteria have to be met, as recommended in the *APA Guidelines for Psychological Assessment and Evaluation* (APA, 2020). Furthermore, regarding the designing of new self-assessment scales for the measurement of interpersonal communication skills, it is known that *LLMs may manifest cultural biases*, as was, for instance, established for ChatGPT (Yuan et al., 2024). In fact, *cultural differences and bias* have been examined in personality assessment for decades (for relevant issues see: van de Vijver and Tanzer,

1997; Heine and Buchtel, 2009; He and van de Vijver, 2017; Dong and Dumas, 2020) and may also be present in interpersonal skills scales generated by LLMs. Finally, despite some optimistic expectations regarding the use of LLMs for the *automation of psychotherapy* (Stade et al., 2024), the training of the more advanced LLMs in this study (i.e. GPT-4o, Gemini 1.5 Pro, and Claude 3.5 Sonnet) for interaction with their users may interfere with such intentions. For example, here is the evaluation that was received by the author of the study that is presented in this paper regarding the results obtained with the 10-item *Active Listening* scale that was, after a single prompt, created, administered, scored and evaluated by one of the LLMs examined: *"Your responses indicate you possess strong active listening skills, which you effectively utilize in your everyday communication. You are clearly a perceptive and empathetic listener, demonstrating a genuine desire to understand others. Your ability to create a comfortable space for open communication, combined with your attention to both verbal and nonverbal cues, allows you to connect with others on a deeper level."* After receiving such favorable feedback from the LLM, the author of this study subjectively concluded that such a 'confident' and 'pleasing' evaluation could have resulted from the LLM's *training for positiveness/appreciation in the conversational style with its users* and, perhaps, even from the intention of the creators of this specific LLM to prepare it (or its successors) to serve as a *supportive personal assistant* in some future commercial application. However, general-purpose LLMs, like those examined in this study, are *not suitable for automated evaluation of scores of communication skills self-assessment scales without considering specific norms for different populations* since their instruction tuning was not performed for *specialized psychological evaluation* based on self-assessment test scores, but more probably for *general-purpose feedback* to better meet their human users' expectations (see: Zhang et al., 2023). As previously illustrated, the utilization of LLMs in automated ('diagnostic') evaluations of users' self-assessments of their interpersonal communication skills should be given proper concern. The potential application of LLMs in the mental health domain, including diagnostics and counseling, currently holds important promises, which are, however, overshadowed by numerous potential perils (Lawrence et al., 2024; Ghuo et al., 2024).

# 8 Conclusion

Even though some reflection upon ethical concerns regarding LLMs have been listed in *subsection 6.6* entitled **"Practical advice regarding the use of LLMs for translation, evaluation, improvement, and design of new self-assessment scales"**, the topic of ethics and LLM (mis)use is so broad and well elaborated in literature (for instance, see: Strasser, 2024; Pournaras, 2023; Chen et al. 2024), including psychology (Abdurahman et al., 2024a) and education (Jiao et al., 2024), that it is not addressed in detail in this paper. However, it is the opinion of the author of this study that researchers should extend and deepen their concerns when considering using LLMs for designing new interpersonal communication skills self-assessment scales. The author of the study presented in this paper decided to *minimize writing* about the design of items for *personality assessment*, except in the brief literature review, since it is a challenge best responded to by *scholars who are experts in that field*. Furthermore, the use of LLMs for questionnaire and survey item generation is also possible in various other fields of research, from healthcare to economics and marketing, but again it may be best for the specialists in those other scientific and applied fields to elaborate on that issue.

Before final comments on the results of the case studies that are presented in this paper, three quotations from interviews with several experts in artificial intelligence ethics are given below, which illustrate the general need for balance between research opportunity and concern in LLM use in research (Birhane et al., 2023): "… the expanding use of these AI technologies in scientific research is the 'discovery of a method of discovery' — the invention of a new set of research tools that support and enable new pathways of insight, innovation and ingenuity in the physical and life sciences."; "A technology that can save time in conducting research and increasing output can be very tempting. But if GenAI is used automatically and without critical oversight, it may fundamentally undermine the foundations of 'good' science."; "Until more robust and reliable safeguards are in place, the scientific community should take a timely and firm stance to avoid any overreliance on LLMs and to foster practices of responsible science in the age of LLMs."

Generative artificial intelligence (GAI), with its widely accessible large language models, like GPT-4o/GPT-4, Copilot, Gemini 1.5 Pro, and Claude 3.5 Sonnet, as their currently most advanced forms, is considered *transformative* (Petroşanu et al., 2023; Sedkaoui and Benaichouba, 2024) and *disruptive technologies*, for instance in the academia (Wong, 2024; Alier et al., 2024) and software development (Sauvola et al., 2024). This transformative and disruptive potential of GAI and LLMs tools is currently also evident in the *specific field of design and evaluation of self-assessment scales and questionnaires for the measurement of interpersonal communication skills*, as was demonstrated in the previous sections of this paper. The ability of publicly accessible LLMs to create items, scales, and questionnaires regarding numerous related constructs in a matter of minutes (even though such instruments created with AI must be subsequently submitted to evaluation by experts and thorough empirical research before responsible practical use on humans) is definitely altering the context of this narrow research environment.

The possibility to use computer-based systems like GPT-4o/GPT-4, Copilot, Gemini 1.5 Pro, and Claude 3.5 Sonnet for the *translation of assessment scale items* is not a novelty since machine translation has been advancing for more than 50 years (Wang et al., 2022). However, what is relevant in the case study (*subsection 6.1*) that is exhibited in this paper is that (a) more than one LLM can be easily accessed by researchers to *generate slightly different versions of translations* of items with the possibility to choose the best one among the provided translations, (b) with the use of more than one LLM, *back-translations* are quickly and easily performed with the use of different LLMs, (c) *fine-tuning of the wording* of translated items is possible with specifically designed prompts, and (d) with the use of LLMs the researchers can *shorten their translations* of items to English and make their wording appear more *natural* in the target language if the original translation was too literal.

The *categorization of items* in initial versions of self-assessment scales and questionnaires (or in item pools) is an important process in the empirical evaluation of their structure and can be performed by *exploratory* and/or *confirmatory factor analysis*. Some of the goals of confirmatory factor analysis (CFA) can be (as adapted from DiStefano and Hess, 2005): (a) develop an initial categorization of constructs; (b) refine an existing theoretical model, (c) confirm the current theoretical structure, (d) test the known theoretical and/or empirical structure in additional populations or contexts, (e) search for alternative conceptualizations of a theory or structural model instead uncritically accepting the 'most obvious' or most easily explained model, (f) investigate 'equivalent' models, (g) search for additional evidence for a model structure by diverse methods and in different populations (gender, age, culture). While CFA commonly implies the use of *empirical data* and exploratory and/or confirmatory factor analyses, as well as SEM, the previously listed potential goals (a-f) can also be at least partly realized by adequate use of *non-empirical semantic analyses* with LLMs. In a study that illustrated various potential uses of LLMs in psychometrics, Franco-Martínez et al. (2023) indicated the potential of LLMs for what they labeled a "more exploratory approach" and "more confirmatory approach" in analyses of research constructs that are related to personality traits. Similarly, in our case studies (presented in subsections *6.2* and *6.3*) the possibility to use specific LLM prompts for 'exploratory semantic analyses' and 'confirmatory semantic analyses' of multifaceted questionnaires and their constituent scale items is demonstrated that can complement the realization of the aforementioned CFA goals. This type of semantic analysis of constructs is an important novelty for interpersonal communication skills assessment that may assist in the improvement of existing self-assessment questionnaires and the construction of new multifaceted measures. In that respect, it must be emphasized that even though in one of the most comprehensive investigations of the skills related to interpersonal communication competence, conducted more than three decades ago, Spitzberg and Cupach (1989) listed about a hundred distinct labels derived from factor analyses of interpersonal variables, a well-organized hierarchical structure or dimensionality (like the Big Five model in personality research) has not yet been widely accepted in this field. In fact, from the latest analysis by Spitzberg (2015) of the dimensionality problem related to various aspects of interpersonal interaction, it can be concluded that numerous approaches and models still exist in parallel without any consensus regarding those that would be most relevant. According to Spitzberg (2015), there are "hundreds of models and conceptual schemes speculating on the structure of interpersonal relations and communicative abilities." Perhaps semantic analyses with the use of LLMs, like the ones that were outlined in the case studies in this paper, could contribute to the organization of both the interpersonal skills-related constructs and the labels for their potential higher-order dimensions.

As was mentioned earlier, the potential to improve the items of self-assessment scales for the measurement of personality traits was demonstrated and discussed by Franco-Martínez et al. (2023). Using adequate prompting, they revealed that GPT-4 can be used to (a) detect wording mistakes in scale items and suggest improvements after it was provided with specific guidance, (b) assess construct representation by individual scale items, (c) assess the dimensionality of both short and long multifaceted personality questionnaires, (d) identify possible constructs that are measured by a given set of items from personality traits assessment scales and make distinctions between items that constitute different measures. Regarding the use of GPT-4 to improve scale items, the authors of this study concluded the following: "... we must be wary that its responses will likely differ from those coming from humans and must be seen as complementary information to address item quality based on theory."; "Indeed, the LLM is capable of identifying theoretically relevant response processes related to item contents and could serve as a means to widen our knowledge on non-cognitive item generation." In *subsection 6.3* of the case study presented here in this paper, a demonstration is provided with the items and scales that have not been previously published, which avoids a possible methodological issue of the study performed by Franco-Martínez et al. (2023), who used the well-known Rosenberg's Self-Esteem Scale (Rosenberg, 1965), which is mentioned in more than 100,000 documents according to the Google Scholar search engine (many of which likely constituted training data for the LLM GPT-4 that they tested). As was demonstrated in the case study in *subsection 6.3* of this paper, when confronted with a multifaceted questionnaire containing a large number of scales and items that were not published and therefore not a part of the training data, an LLM may not perform classification tasks in accordance with the theoretical perspectives of the questionnaire author(s). Also, when given a prompt to perform a meaningless or dubious task, an LLM may uncritically proceed with responding and generate output that resembles 'hallucination' or 'confabulation'. The previously mentioned issues should be a source of concern for the researchers. However, even though this was not thoroughly tested in this study, it is probable that, for questionnaires with a smaller number of constituent scales and items, advanced LLMs may produce apparently meaningful categorizations and observations that could be worth considering and reflecting upon by authors of scales and researchers. Various questions and requirements concerning items, scales, constructs, as well as the entire multifaceted questionnaires, can be posed in well-formulated prompts to LLMs that could in turn provide (a) support for existing conceptualizations, (b) novel critical insights, and (c) novel ideas regarding potential improvements of interpersonal communication skills self-assessment measure.

The potential to design new items for personality assessment measures with the use of natural language processing (NLP) and LLMs has already been demonstrated using a pre-training methodology with GPT-2 (Hernandez et al., 2022; Götz et al., 2023) and GPT-3 (Lee et al., 2023). The use of the general-purpose ChatGPT for the generation of personality items without specific pretraining for the task was described in Walton and Anguiano-Carrasco (2024). Therefore, the automated generation of new self-assessment scales for the *measurement of interpersonal communication skills* does not represent a novelty from a technical perspective, even though the literature search has not revealed scholarly articles dealing with this specific topic. However, the case study that is described in subsection *6.4* of this paper shows that *five publicly available and advanced LLMs* (GPT-4o, GPT-4, Copilot, Gemini 1.5 Pro, and Claude 3.5 Sonnet) can be *effectively used for automated item and scale generation* which was, at the time of this study (June-August 2024), an unprecedented opportunity for its author/researcher.

In terms of demonstration of their potential, all of the examined LLMs were capable with the use of zero-shot prompting to create self-assessment scales for the measurement of five IPIP constructs related to interpersonal interaction (*Compassion*, *Fairness*, *Politeness*, *Impression Management*, and *Enthusiasm*). Also, the LLMs were able to produce definitions of constructs, respective new scale items, as well as the explanation of the new interpersonal communication skills items for the scales labeled *Initiation of Interaction*, *Adaptability in Communication*, *Interaction Involvement*, *Verbal Decoding of Messages*, *Nonverbal Sensitivity*. Furthermore, in section *8* entitled **"Summary and Discussion"** and *Appendix IV* the *contextualization* of the items of the previously mentioned scales from *face-to-face* context to the *Facebook communication* environment is illustrated. Finally, the process of automated generation of the items for a new scale was demonstrated with the use of exemplary items for the *Active Listening* scale. Then again, it has to be admitted that the extraordinary possibility to almost effortlessly create new interpersonal communication skills related items (with multiple LLMs!) imminently raises concern that it may give rise to a plethora of novel scales and questionnaires, beyond the already designed and published instruments, without a quality theoretical background and proper empirical evaluation.

The *multiturn capabilities* of LLMs have not yet been fully explored. For instance, even though LLMs have been frequently investigated in relation to their capability to *solve knowledge and cognitive tests*, and recently as tools for *automated generation of multiple-choice questions* (Doughty et al., 2024; Grévisse et al., 2024; Sayed et al., 2024), the literature search did not yield any research papers that would address their multiturn ability to generate, administer and score individual items, one by one, of a knowledge or personality test. Recent comparisons of LLMs have indicated that GPT-4 and GPT-4o are among the leading LLMs considering *structured output* benchmarks (Liu et al., 2024) and *multiturn conversational abilities* (Kwan et al, 2024). In the case study that is presented in *subsection 6.4* of this paper such multiturn capabilities of various LLMs (GPT-4o, GPT-4, Gemini 1.5 Pro, Claude 3.5 Sonnet) were tested using single zero-shot prompts to both automatically generate a self-assessment interpersonal communication skill scale, and to administer to the user and score the items of the generated scale one by one, while also providing the interpretation of the total score in the end. This was perhaps the first demonstration of its kind regarding this specific LLM capacity reported in the literature, but it must be noted that, in principle, such automated application of diagnostic instruments, especially with the interpretation of the total score or questionnaire score(s) to the user by a general purpose LLM could be ethically questionable.

It can be expected that the multiturn capabilities of LLMs will greatly advance in the future. In fact, in *Appendix V* an example is displayed of a *single prompt* that was successfully used to initiate the following sequence of *automated structured multiturn responses* of GPT-4o and Claude 3.5 Sonnet: (a) creation of a self-assessment interpersonal communication scale to measure 'attentiveness in receiving Facebook messages' using 10 items and a 1 1-5 Likert-type scale; (b) one-by-one administration of the 10 generated items and waiting for the response of the user on the Likert scale; (c) calculating the total score of the user in the scale; (d) performing a Socratic dialogue with 10 question-and-answer multiturn interactions with the user to teach the user about the skill of attentiveness in receiving Facebook messages; (e) generating 10 multiple choice knowledge questions based on the content of the previous Socratic dialogue and asking the user those questions one by one, while also scoring the user's responses in the end; (f) performing the final evaluation including the user's score in the questionnaire, the result of the Socratic dialogue, and the result in the 10 questions of the multiple choice test. The above list of multiturn capabilities of the examined LLMs, initiated by a single prompt, aims to illustrate that researchers may encounter unexpected new capabilities of LLMs and that the specific field of LLM use in designing interpersonal communication skill measures is rapidly progressing.

As demonstrated by the several brief case studies with the currently most advanced LLMs (*GPT-4o*, *GPT-4*, *Copilot*, *Gemini 1.5 Pro*, and *Claude 3.5 Sonnet*) in this research paper, they can be productively used for the specific research purposes of evaluation, improvement, and design of self-assessment scales and questionnaires for the measurement of interpersonal communication skills. Despite the considerable ethical concerns that need to be acknowledged in the design and reporting of the research on the issues that were elaborated in this paper, advanced LLMs, if properly employed, can undeniably facilitate and accelerate scientific advancement in this narrow field.

# 9 Limitations of research

The most important limitation of the research presented in this paper, as well as of other cited research that is published in preprint archives (arXive, SSRN, etc.), is that it has not been peer-reviewed. Only for some of the cited research papers published on preprint servers the information about the conference at which they were listed in the program has been obtained and provided in the list of references, while for other cited preprint papers there is no evidence of peer review.

As was the case with several studies that addressed the early versions of LLMs like GPT-2, GPT-3, ChatGPT, and Bard, novel LLM technologies (GPT-4o, Gemini 1.5 Pro, Claude 3.5 Sonnet) tend to appear in a year or two or less, outperforming the existing ones. Therefore, with the introduction of more advanced LLMs some of the case studies, examples, observations and conclusions that are presented in this research paper may cease to be up-to-date. Also, new research papers are being rapidly produced with findings that may also outdate some of the LLM-related papers that were cited here, even though most of these were published from 2023 until the end of July 2024. However, the potential of using LLMs in evaluating and improving existing interpersonal communication skills questionnaires will probably continue to be a valuable research and applied topic for some time in the future.

Most of the case studies presented in this paper were conducted using multiple LLMs. Therefore, if specific prompting produced comparable results in these case studies with all the examined LLMs, it can be assumed that such results would be reproducible in a similar form for other researchers using the same or more advanced types of LLMs. In the future, after the LLMs have been further considerably improved, more advanced outcomes may be expected with proper prompting (for instance, such a potential improvement in diagnostic performance on radiology quizzes with the use of more advanced LLMs like GPT-4o, Gemini 1.5 Pro, and Claude 3 Sonnet was recently emphasized by Sonoda et al., 2024). It is common knowledge that an LLM may produce (slightly) different responses to the same prompt on various occasions (i.e. different prompt runs). It should thus not be expected that rerunning specific prompts created for the purpose of this study will return exactly the same wording of LLM responses as that presented in the case studies in this paper.

The capacity of the LLMs may vary when the use of a free account is compared with a paid subscription. Also, the most advanced LLMs like GPT-4o are being continuously improved. In that respect, it must be noted that, according to the *Chatbot Arena* (2024), the best-performing LLMs (GPT-4o, Gemini 1.5 Pro, and Claude 3.5 Sonnet) at the beginning of this study (mid-June 2024) were also the best-rated LLMs at the time of its completion (end of August 2024).

Another potentially important concern with the use of LLMs for 'new' content creation in scientific research is the copyright issue. The problem of balancing (a) innovation that may result from the use of LLMs with (b) copyright ownership of the authors of training data for LLMs is indeed of notable complexity (for instance, see: Tyagi, 2024).

When generating new items for interpersonal communication skills scales, most of the LLMs do not automatically provide information on potential sources for their output such as publications and authors of similar items and scales. The 'newly' generated items by the LLMs can be (1) *literal copies*, (2) *non-literal* copies, (3) *alterations* of the original work with *new expressions and meaning*, or (4) *completely original* products (compare: Chen et al., 2024). Researchers should therefore recognize that there may be limitations in the publishing and distribution of newly created scale items by LLMs beyond the *fair usage* principle, especially if the wording of those new items closely resembles or repeats the phrasing of items in the original measures that may have been used to train the LLMs (a similar issue is addressed in: Karamolegkou et al., 2023). To avoid copyright issues regarding the case studies that were presented in this paper (a) the items from the previously unpublished interpersonal communication skills self-assessment scales and questionnaires developed by the author were used (see the instruments in *subsection 5.1.2* entitled *"Self-assessment scales and questionnaires"*) and (b) only selected sample new items produced by each of the LLMs used in this study were revealed for illustration purposes (in *subsection 6.4* named *"Design of new self-assessment scales with LLMs"*).

The capacity of the LLMs to produce a larger number of items and self-assessment scales in zero-shot prompts beyond the number of the newly generated items and scales included in the presented case studies had not been defined as a research question. It was therefore not investigated in this paper.

There are considerable limitations of this study concerning the translation examples. The examined LLMs were not trained in all of the world's languages, but only a small fraction. Also, LLMs demonstrate various abilities to translate from one language to another depending on the amount of training data (*high-resource* languages like English vs. *low-resource* languages) and other factors related to the translation of language pairs (for more information on this issue see for instance: Hendy et al, 2023; Court and Elsner, 2024).

The results in this study apply to general-purpose publicly available LLMs without their fine-tuning (e.g. GPT-4o, GPT-4, Copilot, Gemini 1.5 Pro, and Claude 3.5 Sonnet) and may not correspond to outcomes that may be achieved in similar research with advanced fine-tuned models (for instance, Meta's Llama 3 models; see: Dubey et al., 2024).

# 10 Declaration of conflict of interest and use of AI in research

The author declares that there is no conflict of interest and that (a) this research was not funded by any AI-related project, public or private enterprise, and (b) the research was conducted as part of the authors' scientific research obligation associated with employment at a higher education institution.

Each instance of AI usage in the case studies is denoted in the text of this paper (with a clear indication of the LLM employed for a particular case study). Also, AI was *not used* for literature search, writing of any part of the text of this paper, or for its translation, except for the translation from Croatian to English of the items displayed in *Table 1* and *Table 2*, where using LLMs for translation was associated with the corresponding research question.

# Appendix I

## Sample items of the Interpersonal Communication Competence Inventory – *ICCI*

## (Bubaš, 2003)

*Notes: the original ICCI instrument was developed in 1999 in the Croatian language to investigate the dimensions of interpersonal communication competence and was not designed for diagnostic purposes or other uses; the responses to the items were given on a 1-5 Likert-type scale (1 = Totally untrue, 2 = Mostly untrue, 3 = Neither true nor untrue, 4 = Mostly true, 5 = Totally true); the original items were translated from Croatian to English for the purpose of this study by a non-expert human translator; the ICCI items were not published in their Croatian or English form and detailed construct validation of the ICCI scales was not performed, therefore selected ICCI scales can be used to demonstrate possibilities of improvements of interpersonal communication skills scales by using LLMs; in the continuation, each of the scale labels is followed with a sample of one positively (+) and one negatively worded item (−);*

**Knowledge of The Communication Process**
(+) I am knowledgeable of various skills that contribute to effective communication.
(−) I am uninformed of the rules that govern the behavior of different social communities, structures, or groups.

**Motivation for Interpersonal Communication**
(+) I seek opportunities to socialize with other people.
(−) I'm not willing to socialize a lot or participate in social activities.

**Initiation of Interaction**
(+) Most often, I would be the first person to present himself/herself to someone new in a group of people.
(−) It happens that for days I come close to a lot of people without exchanging a single word with them.

**Communication Effectiveness**
(+) The results of my companionships are the same or better than those of other people.
(−) It is hard for me to make my way in some social environment.

**Assertiveness**
(+) I try to accomplish my goals without significantly jeopardizing the interests of other people.
(−) It is difficult for me to stand up for myself to the extent it is necessary for me.

**Interpersonal Control**
(+) With diverse tactics or strategies, I unobtrusively make people do what suits me best.
(−) I can't handle people with a difficult character.

**Interaction Management**
(+) By a good arrangement of words and actions, I succeed in increasing the effect of my contact with other people.
(−) I frequently fail to discover the right approach to a person that I am contacting.

**Verbal Decoding Skills**
(+) By listening and observing others I can conclude many things about their unspoken attitudes, as well as about what they are prepared to do.
(−) It frequently occurs that I miscomprehend the words or intentions of other people.

**Nonverbal Sensitivity**
(+) Other people find it difficult to conceal what they really feel when I am able to observe them.
(−) Rather often I misinterpret someone's smile.

**Adaptability**
(+) Sudden changes in my social surrounding have never left me without a proper response.
(−) It is hard for me to be a leader in one situation, and a follower in another.

**Altercentrism *(Other-orientedness)***
(+) I like to achieve that others feel content from socializing with me.
(−) I believe that I am generally indifferent toward other people.

**Composure**
(+) At no time does my face "turn to ice" in unexpected or unpleasant situations.
(−) Sometimes I can't manage to produce a response to some question or remark.

**Self-disclosure**
(+) I like to speak with others about true feelings and confidential (intimate) experiences.
(−) I believe that my inner world is not interesting or comprehensible to other people, therefore I do not disclose it to them.

**Empathy**
(+) I can easily imagine myself in other people's shoes, i.e. in the situation of another person.
(−) It is difficult for me to find an explanation for the diverse kinds of behaviors and experiences of other people.

**Comforting**
(+) Even in a hopeless situation I will find something positive and reassuring for the discouraged person.
(−) Others can notice that I regret the time that I spend helping them solve their problems.

**Encoding Skills**
(+) Other people rarely misunderstand my words.
(−) I have a problem when I want to clearly and comprehensively express some complex thought.

**Nonverbal Expressivity**
(+) With my gestures and posture during a speech, I can successfully attract the attention of others to my messages.
(−) I am not prone to excessive gesticulation even when I am with very lively company.

**Verbal Expressivity**
(+) Only with great difficulty would I be able to describe a painting or musical composition.
(−) Even in case of very exciting happenings, I will describe them rather monotonously and indifferently.

**Interaction Involvement**
(+) I get fully engaged in contact with other people and my thoughts don't drift to unrelated themes.
(−) Questions or gestures of other people can find me unprepared for a prompt and suitable response.

**Cooperation**
(+) I am prepared to accept and provide a critique that is placed with good intentions when it helps the achievement of mutual goals.
(−) I behave kindly and helpfully only when it is needed for the realization of my personal aims.

**Social Relaxation**
(+) In the company of a large number of people, I feel at ease and light-hearted.
(−) I seldom feel enough self-assurance to engage in a discussion even about familiar topics.

**Self-monitoring**
(+) I am particularly directed toward monitoring what goes on among people in a certain group so that I can approach them in a suitable manner.
(−) With my uncontrolled facial expressions or posture, I often create a disadvantageous impression.

**Conversational Skill**
(+) People commonly find that conversations with me are very pleasant.
(−) I seldom make an effort to adjust the theme and the flow of a conversation with the opinions and feelings of other speakers.

# Appendix II

## Items of the *joint* scale entitled "Interpersonal Influence and Interaction Management"

Notes: In the table below, the selected 9 out of 24 positively worded items (a translation from Croatian to English) from the original ICCI scales *Interaction Management* (IM), *Assertiveness* (AS), and *Interpersonal Control* (IC) are displayed in the second column and the abbreviations pertaining to each scale in the first column; the data were collected in 1999, N=403, subjects were first-year students of a university study of Information Systems in Croatia, 72% of male and 38% of female gender; the Scree test indicated three factors in the unrotated solution that didn't correspond to the lists of positively worded items of the original scales (IM, AS, IC); Varimax rotation was performed for three fixed factors and factor projections of selected items are presented in the third column in the table below; Cronbach alpha coefficient for the hypothetical self-assessment scale that is provisionally called "Interpersonal Influence and Interaction Management", that would consist of the selected 9 items, was .837; Cronbach's alpha coefficients for this scale if a specific item was deleted, are shown in the fifth column; the semantic analyses performed with GPT-4o indicated the potential to create a single scale ("Interpersonal Influence and Interaction Management") instead of three scales ("Interaction Management", "Assertiveness", and "Interpersonal Control").

| Abbreviation of the ICCI scale | Items of the joint scale "Interpersonal Influence and Interaction Management" | Projection on the first factor | Projections on two other factors | Item-total correlation | Cronbach alpha if the item is deleted |
|---|---|---|---|---|---|
| IM | With my statements and bearing I keep under control the communication in problem situations. | .70 | < .30 | .68 | .807 |
| AS | I can retain my word in a group discussion and thoroughly express my proposal. | .67 | < .20 | .58 | .817 |
| AS | My confidence and self-reliance have a convincing effect on others. | .66 | < .20 | .56 | .819 |
| IM | I can manage what happens in a group even when people are moved by emotions. | .61 | < .20 | .55 | .820 |
| IM | Through my tactical and courteous conduct, I attract people to conversations even about those topics that are unpleasant to them. | .61 | < .30 | .59 | .816 |
| IC | I can manage interpersonal relations in a way that prevents the development of quarrels, conflicts, or fights. | .59 | < .30 | .51 | .824 |
| IM | By a good arrangement of words and actions, I succeed in increasing the effect of my contact with other people. | .59 | < .30 | .57 | .819 |
| IC | Other people feel that it suits them best to conform to my opinion and advice. | .50 | < .30 | .45 | .831 |
| IC | I usually make correct assumptions about how my acts that are directed toward other people will conclude. | .50 | < .20 | .44 | .832 |

# Appendix III

## Categorization of items of the "Active Listening" scale by five LLMs

Prompt given to the five LLMs: "<83 words of the definition of active listening from Wikipedia: https://en.wikipedia.org/wiki/Active_listening>. Act as an expert regarding the interpersonal communication skill 'active listening'. Your assignment is to judge 25 items that may be related to this concept/construct. For each item give one of the following evaluations: 1 = Irrelevant <for active listening>; 2 = Important, but not essential <for active listening>; 3 = Essential <for active listening>. Write your evaluation on the left side of each item. The 25 items to evaluate are: '1. I indicate that I am listening by head nods and appropriate facial expressions.'; <following is a list of the remaining 24 items from the table below>"

Notes: the categories "1 = Irrelevant", "2 = Important, but not essential", and "3 = Essential" were used according to Lawshe (1975); the responses of LLMs (1-3) are presented in respective columns below the names of particular LLMs; the percentage of agreement between the five LLMs regarding particular scale items is displayed in the first column; the items were ranked according to the total (sum) of the individual evaluations by each of the five LLMs; the totals of LLMs' responses could represent the semantic relation of each item with the concept/construct of active listening (as such, this procedure represents a form of Q-sorting analysis); similar analysis can be performed for scales that measure other types of communication skill; it can be observed that all of the items with low total evaluation (7-8) by the five LLMs are *negatively worded* (and, consequently, scored in reverse; these items are written in italics in the following table), as well as that they are semantically related to active listening which makes the categorization of at least some of those items as "1 – Irrelevant" incorrect; with regard to the above, different types of categorization (i.e. different categories) and improved prompting that takes into account the negatively worded items) is advised for improvement of the output of LLMs; it is also possible that the LLMs have used the category "1 = Irrelevant" because this category was included as an option in the prompt without specifically mentioning that some of the categories may not apply to any of the items; this example illustrates the need for precise prompting for achieving desired output from the LLMs.

| Percent of agreement | Total (sorted) | GPT-4o | GPT-4 | Copilot | Gemini 1.5 Pro | Claude 3.5 Sonnet | Item no. | Items from the self-assessment scale "Active Listening" |
|---|---|---|---|---|---|---|---|---|
| 100% | 15 | 3 | 3 | 3 | 3 | 3 | 1. | I indicate that I am listening by head nods and appropriate facial expressions. |
| 100% | 15 | 3 | 3 | 3 | 3 | 3 | 3. | In face-to-face conversations, I maintain good eye contact. |
| 100% | 15 | 3 | 3 | 3 | 3 | 3 | 17. | If someone expresses his/her thoughts or ideas poorly or unclearly, I still make an effort to listen to what this person wishes to say. |
| 100% | 15 | 3 | 3 | 3 | 3 | 3 | 21. | I am cautious not to omit something when others are talking to me, and I ask questions to acquire complete information. |
| 100% | 15 | 3 | 3 | 3 | 3 | 3 | 24. | I make an effort to put together all the details of another person's speech to create an orderly and integral "picture" or conception of his/her message in my mind. |
| 80% | 14 | 3 | 3 | 3 | 3 | 2 | 19. | If a person is unable to articulate an idea, I aid or guide the efforts of this person with consideration. |
| 60% | 13 | 3 | 3 | 3 | 2 | 2 | 8. | I can identify with other people's experiences and feelings even when they are quite different from my own. |
| 60% | 13 | 2 | 2 | 3 | 3 | 3 | 11. | I can unceasingly concentrate on the content of another person's long speech. |
| 60% | 13 | 3 | 2 | 3 | 2 | 3 | 18. | I judge other people's spoken thoughts and opinions independently of their looks or my overall impressions of them. |
| 60% | 12 | 3 | 2 | 2 | 2 | 3 | 4. | I avoid unnecessary movements or activities when others are speaking to me. |

| Percent of agreement | Total (sorted) | GPT-4o | GPT-4 | Copilot | Gemini 1.5 Pro | Claude 3.5 Sonnet | Item no. | Items from the self-assessment scale "Active Listening" |
|---|---|---|---|---|---|---|---|---|
| *Continued from the previous page!* | | | | | | | | |
| 60% | 12 | 3 | 2 | 2 | 2 | 3 | 12. | I make efforts to follow how consistent, reasonable, and substantiated other people's orations are. |
| 80% | 11 | 2 | 2 | 3 | 2 | 2 | 6. | I find it difficult to react in the right way when the person who is talking to me expresses intense sorrow or joy. |
| 80% | 11 | 2 | 2 | 3 | 2 | 2 | 10. | People feel comforted after talking to me about their worries even when we don't solve their problems. |
| 80% | 11 | 2 | 2 | 2 | 2 | 3 | 22. | While listening, I try to distinguish facts from emotions and impressions that are created by the speaker's gestures. |
| 60% | 10 | 2 | 2 | 3 | 1 | 2 | 7. | When friends or colleagues refer to me, I have an understanding of all their problems. |
| 60% | 8 | 2 | 1 | 2 | 1 | 2 | 9. | *I am inhibited from sharing feelings of happiness, worries, or grief with someone else.* |
| 60% | 8 | 2 | 1 | 3 | 1 | 1 | 25. | *After a person I am talking with begins a lengthy speech, I find it increasingly difficult to follow up on all that he/she means to say.* |
| 80% | 7 | 1 | 1 | 3 | 1 | 1 | 2. | *I interrupt others before they finish what they mean to say.* |
| 80% | 7 | 1 | 1 | 3 | 1 | 1 | 5. | *People can notice when I find it dull to listen to what they are telling me.* |
| 80% | 7 | 1 | 1 | 3 | 1 | 1 | 13. | *My thoughts wander off to unrelated topics or focus on something else in my environment when someone is speaking to me.* |
| 80% | 7 | 1 | 1 | 3 | 1 | 1 | 14. | *I am easily distracted by sounds or changes in the surroundings while listening to what others are telling me.* |
| 60% | 7 | 2 | 1 | 2 | 1 | 1 | 15. | *After a discussion, I am unable to correctly and concisely retell what has been said to me.* |
| 80% | 7 | 1 | 1 | 3 | 1 | 1 | 16. | *After realizing that my beliefs are opposite of those of another person, I quickly lose the willingness to give attention to what he/she is telling me.* |
| 80% | 7 | 1 | 1 | 3 | 1 | 1 | 20. | *When I dislike someone, I lack interest in the words and thoughts he/she may try to communicate to me.* |
| 80% | 7 | 1 | 1 | 3 | 1 | 1 | 23. | *I draw conclusions before others have finished what they intended to tell me.* |

# Appendix IV

## Contextualization of the items of the selected ICCI self-assessment scales to Facebook communication

The prompt that was given to the five LLMs for the contextualization task (including the 5 scales and 8 positively worded items for each scale): "Please rephrase the items or create new items for the following skills in face-to-face communication (written in capital letters) so that they can apply to online communication using the Facebook social network. For each of the listed skills, there should be 5 new Facebook-related items. Write the name of each skill followed by 5 new items:

INITIATION OF INTERACTION

01. Most often, I would be the first person to present himself/herself to someone new in a group of people.

02. I commonly gather a larger or a smaller crowd to have a good time, for a party or a celebration.

03. It is easy for me to initiate contact with very different people.

04. It is no problem for me to start familiarizing with a person of a different gender.

05. I could always mingle into an entirely unknown group of people during a celebration or a party.

06. I use every suitable instance or situation for making contact with interesting people.

07. I use diverse techniques or excuses to secure an opportunity for conversation with a person that holds a position of importance to me.

08. I like to call other people to go out together, or to engage in some outdoor activity.

ADAPTABILITY IN COMMUNICATION

01. I am able to quickly find several potentially good responses to some difficult interpersonal situations.

02. I can easily accept different social roles and carry them out successfully.

03. Sudden changes in my social surroundings have never left me without a proper response.

04. The way that I behave and contact others I will always effectively adapt to my important goals.

05. The social position and cultural characteristics of other people commonly influence the approach that I choose, and the manner in which I speak when I address them.

06. I successfully improvise when I don't have in advance a prepared answer to some novel social situation.

07. The bearing and conduct of other people are a good guideline for me how to better adapt my behavior.

08. I can preserve face and ensure the favor and support of others in a situation that is unpleasant for me.

INTERACTION INVOLVEMENT

01. People whom I contact perceive that I am fully dedicated to our shared discussion.

02. am usually more active than others when a certain subject is being discussed in a group.

03. I get fully engaged in contacts with other persons and my thoughts don't drift to unrelated themes.

04. I can react instantaneously and straightforwardly when meeting with other people.

05. The words that I say and how I behave usually has a lot of impact on the atmosphere within an entire group.

06. Almost at any moment I can appropriately take the floor in a discussion.

07. When communicating with several people my attention is constantly focused on them, as well as on my role in mutual contacts.

08. I try to attentively follow the topics in a discussion, as well as the opinions of other speakers.

VERBAL DECODING OF MESSAGES

01. I always make a fair estimate of what is going on among people in some social situation.

02. By listening and observing others I can conclude many things about their unspoken attitudes, as well as about what they are prepared to do.

03. I always take into account many potential causes for someone's behavior.

04. I can easily differentiate indisputable facts from unverified opinions and hasty estimates of other people.

05. I never take other people's messages literally, but I compare them with the situation they are in.

06. I quickly understand what someone means to say, even when he/she is expressing himself/herself unclearly.

07. I can "read between lines" when someone writes me a note, and interpret sudden muteness in a conversation.

08. I manage to determine the key points of every difficult situation that involves other people.

NONVERBAL SENSITIVITY

01. The intensity, tone and trembling of a person's voice tell me much more than his/her words.

02. When someone blushes the exact reason for that is always easily understandable to me.

03. I like to observe other people's bearing and movements because they are a source of most diverse messages.

04. Other people find it difficult to conceal what they really feel when I am able to observe them.

05. Sometimes I am completely focused on monitoring someone's behavior.

06. It is easy for me to notice when someone is upset even though he/she didn't reveal it with a single word.

07. I am never mistaken when I notice a friendly approach of a certain person.

08. I am usually the first one to recognize when someone begins to hold himself/herself rigidly, or when someone's hand or voice trembles."

## Selection of items contextualized to Facebook communication

Notes: the original ICCI items are written in boldface and numbered as they appear in the above prompt; the selected contextualized items are written in italics and the LLMs that produced them are denoted in angled brackets; scale labels are written in uppercase.

### INITIATION OF INTERACTION

**01. Most often, I would be the first person to present himself/herself to someone new in a group of people.**
*<GPT-4o> I often initiate conversations with new people by sending friend requests on Facebook.*
**02. I commonly gather a larger or a smaller crowd to have a good time, for a party or a celebration.**
*<Copilot> Whether it's a virtual party or a celebration, I can easily bring people together on Facebook.*
**03. It is easy for me to initiate contact with very different people.**
*<Claude 3.5 Sonnet> I find it easy to send friend requests or initiate conversations with diverse people on Facebook.*
**05. I could always mingle into an entirely unknown group of people during a celebration or a party.**
*<GPT-4 5> I frequently join new and unfamiliar Facebook groups and actively participate in their discussions.*
**06. I use every suitable instance or situation for making contact with interesting people.**
*<Gemini 1.5 Pro> I use Facebook Messenger to start conversations with people I find interesting, whether it's about their posts, shared interests, or to ask a question.*

### ADAPTABILITY IN COMMUNICATION

**01. I am able to quickly find several potentially good responses to some difficult interpersonal situation.**
*<Claude 3.5 Sonnet> I can quickly think of multiple appropriate responses to challenging situations in Facebook comments.*
**02. I can easily accept different social roles and carry them out successfully.**
*<Copilot> Adapting to different roles (from casual friend to professional contact) is second nature to me on Facebook.*
**03. Sudden changes in my social surroundings have never left me without a proper response.**
*<Gemini 1.5 Pro> I can seamlessly move from public posts and comments to private messages or group chats as the situation demands.*
**05. The social position and cultural characteristics of other people commonly influence the approach that I choose, and the manner in which I speak when I address them.**
*<GPT-4> I adjust my communication style on Facebook based on the cultural background and social status of my contacts.*
**07. The bearing and conduct of other people are a good guideline for me how to better adapt my behavior.**
*<GPT-4o> The behavior and communication style of others provide me with good guidelines to better adapt my online responses.*

### INTERACTION INVOLVEMENT

**01. People whom I contact perceive that I am fully dedicated to our shared discussion.**
*<Claude 3.5 Sonnet> My Facebook friends notice that I'm fully engaged in our conversations through comments and Messenger.*
**02. I am usually more active than others when a certain subject is being discussed in a group.**
*<GPT-4> I am often more active than others in Facebook group discussions, especially on topics I'm passionate about.*
**03. I get fully engaged in contacts with other persons and my thoughts don't drift to unrelated themes.**
*<Gemini 1.5 Pro> I keep my comments and responses relevant to the original post or discussion thread on Facebook.*
**04. I can react instantaneously and straightforwardly when meeting with other people.**
*<Copilot 4> Whether it's a message or a comment, I respond promptly on Facebook.*
**08. I try to attentively follow the topics in a discussion, as well as the opinions of other speakers.**
*<GPT-4o> I try to attentively follow discussion threads and the opinions of other participants on Facebook.*

### VERBAL DECODING OF MESSAGES

**02. By listening and observing others I can conclude many things about their unspoken attitudes, as well as about what they are prepared to do.**
*<Gemini 1.5 Pro> I pay close attention to the wording, punctuation, and emojis used in Facebook messages to accurately interpret the sender's tone and intent.*
**03. I always take into account many potential causes for someone's behavior.**
*<GPT-4> I consider various potential reasons behind people's posts and comments on Facebook before responding.*
**04. I can easily differentiate indisputable facts from unverified opinions and hasty estimates of other people.**
*<Claude 3.5 Sonnet> I can distinguish between factual information and opinions in Facebook posts and discussions.*
**05. I never take other people's messages literally, but I compare them with the situation they are in.**
*<Copilot> Rather than taking things literally, I analyze Facebook messages within their context.*
**07. I can "read between lines" when someone writes me a note, and interpret sudden muteness in a conversation.**
*<GPT-4o> 7. I can "read between the lines" in Facebook messages and interpret sudden silences in conversations.*

### NONVERBAL SENSITIVITY

**01. The intensity, tone and trembling of a person's voice tell me much more than his/her words.**
*<GPT-4o> The choice of words, emojis, and typing style in Facebook messages tell me much more than the literal content.*
**02. When someone blushes the exact reason for that is always easily understandable to me.**
*<GPT-4> I can often understand the reasons behind someone's reactions on Facebook, such as likes, angry reacts, or sad reacts.*
**03. I like to observe other people's bearing and movements because they are a source of most diverse messages.**
*<Claude 3.5 Sonnet> I observe people's profile pictures, cover photos, and shared content as sources of nonverbal cues.*
**04. Other people find it difficult to conceal what they really feel when I am able to observe them.**
*<Copilot 4.> Facebook friends can't hide their true emotions—I notice subtle signs.*
**05. Sometimes I am completely focused on monitoring someone's behavior.**
*<Gemini 1.5 Pro> I pay attention to the specific emojis and reactions used on Facebook to gain further insight into the sender's emotions and intentions.*

# Appendix V

## Illustration of a single one-shot prompt that tests the structured multiturn response capability of advanced LLMs

Here is an example of a single prompt that initiates a sequence of automated and structured multiturn responses (it was successfully tested with GPT-4o and Claude 3.5 Sonnet): "Create a scale with 10 self-assessment items related to attentiveness in receiving Facebook messages, with a Likert-type response scale ranging from '1 = Not at all true of me' to '5 = Totally true of me' and ask me one by one item, waiting for my response to each item. I repeat, ask me one by one item and wait for my response. After you get all my responses, calculate the total score. Then perform a Socratic dialogue with me by which you will try to teach me about attentiveness in Facebook communication. Use a maximum of 10 turns with your questions and my answers. After completing the Socratic dialogue, using the broad content of your questions and my responses in that dialogue, create a knowledge test with 10 multiple choice questions with four possible responses to each, marked A), B), C), and D), of which only one response is correct. Ask me one by one of those 10 questions in the test and wait for my response with 'A', 'B', 'C' or 'D'. Score my result on this test. Finally, give me feedback on the level of my skill of attentiveness in receiving Facebook messages, the results of my learning with the Socratic method, and the knowledge I have demonstrated with my responses to the test questions."